\documentclass[12pt]{article}
\usepackage{natbib}
\usepackage{amsmath}
\usepackage{mathtools}
\usepackage{xfrac}
\usepackage{psfrag,epsf}
\usepackage{float}
\usepackage{colortbl}

\usepackage{graphicx}
\usepackage{enumerate}
\usepackage{url} 
\usepackage{amssymb}

   \usepackage[compact]{titlesec}
    \titlespacing{\section}{0pt}{2ex}{1ex}
    \titlespacing{\subsection}{0pt}{1ex}{0ex}
    \titlespacing{\subsubsection}{0pt}{0.5ex}{0ex}

\usepackage{rotating}

\usepackage{hyperref}

\newcommand{\blind}{1}

\DeclareMathOperator{\E}{\mathbb{E}}
\DeclarePairedDelimiter\abs{\lvert}{\rvert}%
\DeclarePairedDelimiter\norm{\lVert}{\rVert}%

\begin{document}

\def\spacingset#1{\renewcommand{\baselinestretch}%
{#1}\small\normalsize} \spacingset{1}

\def\bSig\mathbf{\Sigma}

\newcommand{\VS}{V\&S}
\newcommand{\tr}{\mbox{tr}}
\newcommand{\pkg}[1]{\texttt{#1}}

\makeatletter
\let\oldabs\abs
\def\abs{\@ifstar{\oldabs}{\oldabs*}}
\let\oldnorm\norm
\def\norm{\@ifstar{\oldnorm}{\oldnorm*}}
\makeatother

\if1\blind
{
  \title{\bf Deeply-Learned Generalized Linear Models with Missing Data}
  \author{David K. Lim\hspace{.05cm}\thanks{The authors gratefully acknowledge NIH grants U01-CA274298, P50-CA257911, P50-CA058223, T32-CA106209, 1R01AA02687901A1, and 1OT2OD032581-02-321, and NSF grants IIS2133595 and DMS2324394 for funding this research.}\hspace{.2cm}\\
    Department of Biostatistics, University of North Carolina at Chapel Hill\\
    and \\
    Naim U. Rashid \\
    Department of Biostatistics, University of North Carolina at Chapel Hill \\
    and \\
    Junier B. Oliva \\
    Department of Computer Science, University of North Carolina at Chapel Hill \\
    and \\
    Joseph G. Ibrahim \\
    Department of Biostatistics, University of North Carolina at Chapel Hill}
  \maketitle
} \fi

\if0\blind
{
  \bigskip
  \bigskip
  \bigskip
  \begin{center}
    {\LARGE\bf Deeply-Learned Generalized Linear Models with Missing Data}
\end{center}
  \medskip
} \fi

\bigskip
\begin{abstract}

Deep Learning (DL) methods have dramatically increased in popularity in recent years, with significant growth in their application to various supervised learning problems. However, the greater prevalence and complexity of missing data in such datasets present significant challenges for DL methods. Here, we provide a formal treatment of missing data in the context of deeply learned generalized linear models, a supervised DL architecture for regression and classification problems. We propose a new architecture, \textit{dlglm}, that is one of the first to be able to flexibly account for both ignorable and non-ignorable patterns of missingness in input features and response at training time. We demonstrate through statistical simulation that our method outperforms existing approaches for supervised learning tasks in the presence of missing not at random (MNAR) missingness. We conclude with a case study of the Bank Marketing dataset from the UCI Machine Learning Repository, in which we predict whether clients subscribed to a product based on phone survey data. Supplementary materials for this article are available online.

\end{abstract}

\noindent%
\vfill

\newpage
\spacingset{1.5} 


\section{Introduction} \label{sec:intro}

Deep Learning (DL) methods have been increasingly used in an array of supervised learning problems in various fields, for example, in the biomedical sciences \citep{Razzak2017,Lopez2018}. While a number of deep learning architectures have been proposed for supervised learning, the feed forward neural network (FFNN) is commonly used in most architectures. In a FFNN, sequential non-linear transformations are applied to the values of the input layer. Each value in the subsequent layer of the FFNN is computed by applying a non-linear (or ``activation'') function to the linear transformation of the values in the previous layer, outputting a complex non-linear transformation of the input \citep{Svozil1997}. For example, a FFNN architecture called the deeply-learned GLM \citep{Tran2019} has been applied in the context of supervised learning to describe nonlinear relationships between the covariates and the response. Due to the large number of parameters, so optimization is often done via stochastic gradient descent \citep{Guo1990}.

However, the common presence of missing data in datasets can hinder the training and generalizability of supervised deep learning methods \citep{Wells2013}, where missingness can occur both in the input features and the response variable. Missingness has commonly been categorized into three mechanisms: Missing Completely At Random (MCAR), Missing At Random (MAR), and Missing Not At Random (MNAR) \citep{Rubin1976}. While a number of methods have been proposed in the statistical literature to address MNAR missingness in the regression setting \citep{Ibrahim2005}, such methods often cannot take into account complex relationships between predictors and response and are not scalable to higher dimensions \citep{Chen2019}, or have been specifically designed for unsupervised learning tasks \citep{Lim2021}. Supervised deep learning is one way to address capture complex relationships between predictors and response in a scalable manner \citep{Kingma2019}, however it is unclear how best to account for more complex forms of missingness, such as MAR or MNAR missingness, in this setting.

There have been some recent attempts to perform prediction using deep learning in the presence of missing features \citep{Ipsen2021}, but such methods typically assume either MCAR or MAR missingness. Commonly used missing data methods in supervised deep learning applications, such as mean imputation or complete case analysis, have historically yielded biased results \citep{Ibrahim2009}. Multiple Imputation by Chained Equations (\textit{mice}) has also been widely employed to account for missing data in a supervised learning setting. However, \textit{mice} is unable to apply a trained imputation model to handle missingness that may exist at test time \citep{Hoogland2020}. In addition, multiple imputation-based methods may not be feasible to apply when the downstream model is computationally intensive, such as in the setting of training a deep learning neural network, since one must train the model separately for each imputed dataset. Lastly, existing approaches to handle MAR or MCAR missingness when training deep learning models for supervised learning tasks are currently limited, and have not been sufficiently explored in the literature. 

To address these issues, we present \textit{dlglm}: a deep generalized linear model (GLM) for probabilistic supervised learning in the presence of missing input features and/or response across a variety of missingness patterns. Our proposed method utilizes variational inference to learn approximate posterior distributions for the missing variables, and replaces missing entries with samples from these distributions during maximization. In this way, \textit{dlglm} can perform supervised learning in the presence of missingness in both the features and the response of interest. We also incorporate a model for the missingness, which can take into account MNAR patterns of missingness, even at training time. Through neural networks, \textit{dlglm} is able to model complex non-linear relationships between the input features and the response, and is scalable to large quantities and dimensionalities of data. Prediction can be done seamlessly on fully- or partially-observed samples using the trained model, without requiring separate imputation of the missing values.

%


\section{Methods} \label{sec:meth3}
Here we first discuss the formulation of the generalized linear model (GLM) in Section \ref{sec:glm}, and then introduce the deeply-learned GLM in Section \ref{sec:deepglm}. We then discuss missingness in the context of GLMs in Section \ref{sec:missglm}, and lastly propose a novel deep learning architecture \textit{dlglm} in Section \ref{sec:dlglm} to fit deeply learned GLMs in the presence of missingness.

\subsection{Generalized Linear Models (GLMs)} \label{sec:glm}
Let $\mathbf{X}$ be the $n \times p$ matrix of covariates (input features) with observation vectors $\mathbf{x}_i$, where each corresponding entry $x_{ij}$ denotes the value of the $i^{th}$ observation of the $j^{th}$ feature for $i=1,\ldots,n$ and $j=1,\ldots,p$. Also, let $\mathbf{Y}=\{y_1,\ldots,y_n\}$ be the vector of univariate responses where $y_i$ is the response pertaining to the $i^{th}$ observation.  We note that $y_i$ may also be assumed to be multivariate; however, we focus specifically on the case of univariate response to simplify the discussion, and discuss extensions to the setting of multivariate response in Section \ref{sec:disc3}. Then, denote $\boldsymbol{\eta}=\mathbf{X}\boldsymbol{\beta}$, where $\boldsymbol{\beta}$ is a vector of regression coefficients and $\boldsymbol{\eta}$ is the linear predictor. Also define $\boldsymbol{\mu}=\{\mu_1,\ldots,\mu_n\}$ with $\mu_i=E(y_i|\mathbf{x}_i)$ and link function $g(\cdot)$ such that $g(\mu_i)=\eta_i=\mathbf{x}_i\boldsymbol{\beta}$.  We assume that the conditional distribution $p(y_i|\mathbf{x}_i)$ is a member of the exponential family of distributions \citep{Mccullagh2019}, such that $p(y_i|\mathbf{x}_i)$ can be written as $$p(y_i|\mathbf{x}_i)=\text{exp} \left[ \frac{y_i\Theta_i-b(\Theta_i)}{a(\alpha)} + c(y_i,\alpha) \right] ,$$ with canonical parameter $\Theta_i$,  dispersion parameter $\alpha$, and some functions $a(\cdot)$, $b(\cdot)$, and $c(\cdot)$. Here, we further assume $g(\cdot)$ is a canonical link function such that $g(\mu_i)=\Theta_i$. With the appropriate specification of the canonical link $g(\cdot)$ and variance function $V_\alpha(\cdot)$, we obtain the formulation of a GLM.

GLMs were first motivated by the limitations of the traditional linear model, which imposed strict assumptions of linearity between $\boldsymbol{\mu}$ and $\mathbf{X}$ and of normality of errors with fixed variance. GLMs instead utilize specific link and variance functions, allowing for model fitting on types of response data that may violate these assumptions, such as count or categorical outcomes, without having to rely on heuristic transformations of the data \citep{Nelder1972}. Typically, GLMs are estimated by utilizing iteratively re-weighted least squares in lower dimensions \citep{Holland1977}, with extensions to the higher dimensional case via penalized likelihood \citep{Friedman2010}. 


\subsection{Deeply Learned GLMs} \label{sec:deepglm}
\begin{figure}
\begin{center}
\includegraphics[width=95mm]{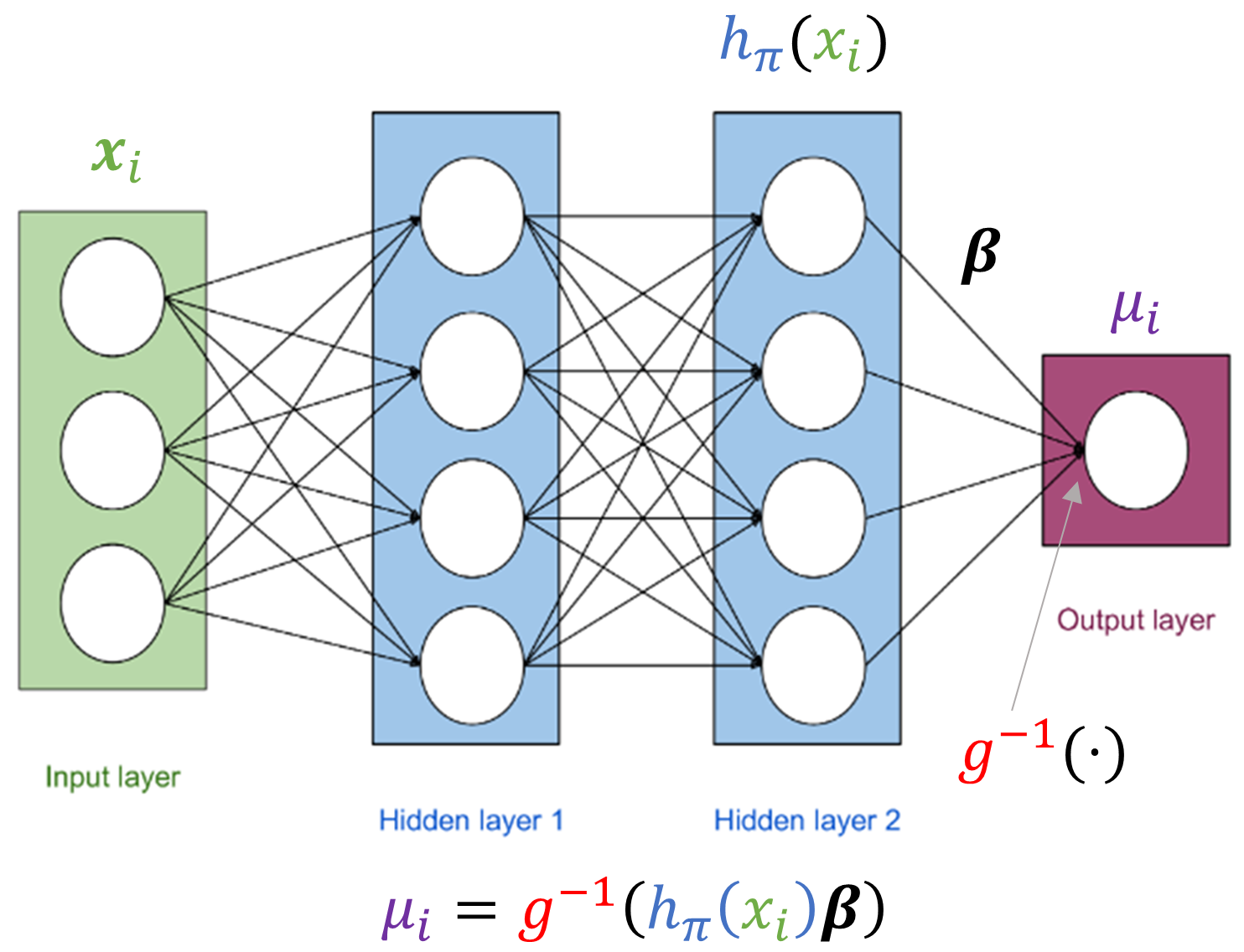}
\end{center}
\caption{\scriptsize Visualization of a sample deeply-learned GLM architecture $s_{\pi,\boldsymbol{\beta}}(\mathbf{x}_i)$. Here, $\pi$ denotes the set of weights and biases pertaining to the portion of the architecture from the input layer to the second to last layer (hidden layer 2). $h_\pi(\mathbf{x}_i)$ is a subset of the entire architecture, such that $s_{\pi,\boldsymbol{\beta}}(\mathbf{x}_i) = h_\pi(\mathbf{x}_i)\boldsymbol{\beta}$. Original artwork of a feed-forward neural network \citep{Dormehl2019} was modified to show deeply-learned GLM architecture.}
\label{fig:ANNGLM}
\end{figure}

The traditional GLM assumes $g(\mu_i)$ is a linear function of $\mathbf{x}_i$, i.e. $g(\mu_i)=\mathbf{x}_i\boldsymbol{\beta}$. In many modern applications, one may wish to model $g(\mu_i)$ as a  non-linear function of $\mathbf{x}_i$ or capture complex interactions between features to predict response \citep{Qi2003}. In such cases, we may generalize the GLM to a deeply-learned GLM \citep{Tran2019} with the following expression: $g(\mu_i) = \eta_i = h_\pi(\mathbf{x}_i)\boldsymbol{\beta}$, where $h_\pi(\cdot)$ denotes the output of a series of non-linear transformations applied to the input $\mathbf{X}$ by a neural network, with weights and bias parameters denoted by $\pi$. In addition, $\eta_i$ can alternatively be expressed  $\eta_i=s_{\pi,\boldsymbol{\beta}}(\mathbf{x}_i)$, where $s_{\pi,\boldsymbol{\beta}}(\cdot)$ is a neural network where $\boldsymbol{\beta}$ denotes the weights and bias associated with the output (last) layer of $s_{\pi,\boldsymbol{\beta}}(\cdot)$. This formulation allows for the traditional interpretation of $\boldsymbol{\beta}$ as the coefficients pertaining to a transformed version of the input covariates. Figure \ref{fig:ANNGLM} shows an illustration of this architecture. 

Let $n_{HL}$ denote the number of hidden layers in $s_{\pi,\boldsymbol{\beta}}(\cdot)$. We note that if $n_{HL}=0$, then  $h_\pi(\mathbf{x}_i)=\mathbf{x}_i$ and $s_{\boldsymbol{\beta}}(\mathbf{x}_i) = \mathbf{x}_i\boldsymbol{\beta}$, reducing to the traditional GLM. Deeply learned GLMs and other neural networks are often maximized using stochastic gradient descent \citep{Bottou2012}. Details of this algorithm can be found in Appendix A1 of the supplementary materials. 


\subsection{Missingness in GLMs} \label{sec:missglm}
Many modern datasets often contain complex forms of missingness \citep{Ghorbani2018}. In GLMs, missingness can exist in either $\textbf{X}$ or $\textbf{Y}$. Therefore, we specify three cases of missingness in this context: missing covariates with fully-observed response (\textit{Case x}), missing response with fully-observed covariates (\textit{Case y}), and missing covariates and missing response (\textit{Case xy}). Define $\mathbf{R}=\{\mathbf{R}^X,\mathbf{R}^Y\}$ as the ``missingness mask'', which denotes the missingness of $\{\mathbf{X},\mathbf{Y}\}$, such that $\mathbf{R}^X$ and $\mathbf{R}^Y$ have the same dimension as $\mathbf{X}$ and $\mathbf{Y}$, respectively, and a value of 1 in $\mathbf{R}$ denotes that the corresponding entry in $\{\mathbf{X},\mathbf{Y}\}$ is observed, while a value of 0 denotes that it is unobserved. Additionally, let $\mathbf{R}=\{\mathbf{r}_1,\ldots,\mathbf{r}_n\}$ with $\mathbf{r}_i = \{\mathbf{r}_i^X, r_i^Y\} = \{r_{i1}^X,\ldots,r_{ip}^X, r_i^Y\}$, $\mathbf{R}^X=\{\mathbf{r}^X_1,\ldots,\mathbf{r}^X_n\}$, and $\mathbf{R}^Y=\{r^Y_1,\ldots,r^Y_n\}$ with elements $\mathbf{r}^X_i$ and $r^Y_i$ pertaining to the missingness of the $i^{th}$ observation of $\mathbf{X}$ and $\mathbf{Y}$, respectively. Then, $\mathbf{X}$ and $\mathbf{Y}$ can be factored into the unobserved and observed entries $\{\mathbf{X}^m,\mathbf{X}^o\}$ and $\{\mathbf{Y}^m,\mathbf{Y}^o\}$, respectively, such that $\mathbf{X}^m=\{\mathbf{X}:\mathbf{R}^X=0\}$ with $\mathbf{x}_i^m=\{\mathbf{x}_i:\mathbf{r}_i^X=0\}$, $\mathbf{X}^o=\{\mathbf{X}:\mathbf{R}^X=1\}$ with $\mathbf{x}_i^o=\{\mathbf{x}_i:\mathbf{r}_i^X=1\}$, and $\mathbf{Y}^m = \{\mathbf{Y}:\mathbf{R}^Y=0\}$ and $\mathbf{Y}^o = \{\mathbf{Y}:\mathbf{R}^Y=1\}$, with $y_i^m=\{y_i: r_i^Y=0\}$ and $y_i^o=\{y_i: r_i^Y=1\}$.


Missingness was classified into three primary mechanisms in the seminal work by \citet{Little2002}: missing completely at random (MCAR), missing at random (MAR), and missing not at random (MNAR). They satisfy the following relations:
\begin{itemize}
\item MCAR: $p(\mathbf{r}_{i}|\mathbf{x}_i,y_i,\boldsymbol{\phi}) = p(\mathbf{r}_i|\boldsymbol{\phi})$
\item MAR: $p(\mathbf{r}_{i}|\mathbf{x}_i,y_i,\boldsymbol{\phi}) = p(\mathbf{r}_i|\mathbf{x}_i^o,y_i^o,\boldsymbol{\phi})$
\item MNAR: $p(\mathbf{r}_{i}|\mathbf{x}_i,y_i,\boldsymbol{\phi}) = p(\mathbf{r}_i|\mathbf{x}_i^o,\mathbf{x}_i^m,y_i^o,y_i^m,\boldsymbol{\phi})$.
\end{itemize}
Here, $\boldsymbol{\phi}$ denotes the collection of parameters for the model of the missingness mask $\mathbf{r}_i$. In the presence of missingness, the marginal log-likelihood can generally be written as
\begin{align*}
\log p_{\alpha, \boldsymbol{\beta}, \pi, \psi, \phi}(\mathbf{X}^o,\mathbf{Y}^o,\mathbf{R})&=\log \iint p_{\alpha, \boldsymbol{\beta}, \pi, \psi, \phi}(\mathbf{X},\mathbf{Y},\mathbf{R}) d\mathbf{X}^m d\mathbf{Y}^m \\
&= \log \iint p_{\alpha,\boldsymbol{\beta}, \pi}(\mathbf{Y}|\mathbf{X})p_\psi(\mathbf{X})p_{\phi}(\mathbf{R}|\mathbf{X},\mathbf{Y}) d\mathbf{X}^m d\mathbf{Y}^m, \stepcounter{equation} \tag{\theequation} \label{eqn:LLNonignorable3}
\end{align*}
where $\psi$ is a set of parameters associated with the covariate distribution $p_\psi(\mathbf{X})$.   We factor $p_{\alpha, \boldsymbol{\beta}, \pi, \psi, \phi}(\mathbf{X},\mathbf{Y},\mathbf{R})$ using the selection model factorization \citep{Diggle1994}. 

Under MNAR, it is not possible to remove $p_{\phi}(\mathbf{R}|\mathbf{X},\mathbf{Y})$ from the integral, since $\mathbf{R}$ can depend on $\{\mathbf{X}^m, \mathbf{Y}^m\}$.  Therefore, MNAR missingness is said to be non-ignorable, because it requires specification of the so-called ``missingness model'' $p(\mathbf{r}_{i}|\mathbf{x}_i,y_i,\boldsymbol{\phi})$ \citep{Stubbendick2003}. There are a number of ways to specify this model. For example, \citet{Diggle1994} proposes a binomial model for the missing data mechanism, which can be written in this setting as $$p(\mathbf{R}|\mathbf{X},\mathbf{Y},\boldsymbol{\phi}_{j_m})=\prod_{i=1}^{n} \prod_{j_m=1}^{p_{miss}} \left[ p(r_{ij_m}=1|\mathbf{x}_i,y_i,\boldsymbol{\phi}_{j_m})\right]^{r_{ij_m}}\left[ 1-p(r_{ij_m}=1|\mathbf{x}_i,y_i,\boldsymbol{\phi}_{j_m})\right]^{1-r_{ij_m}},$$ where $j_m=1,\ldots,p_{miss}$ indexes the $p_{miss}$ features in $\{\mathbf{X},\mathbf{Y}\}$ that contain missingness.  Here $p_{miss} = p_{miss}^X + p_{miss}^Y$, where $p_{miss}^X$ is the total number of features containing missingness in $\mathbf{X}$, and $p_{miss}^Y$ is 1 if $\mathbf{Y}$ contains missingness (0 otherwise). Also, $\boldsymbol{\phi}_{j_m}$ is the set of coefficients pertaining to the missingness model of the $j_m^{th}$ missing variable, and $p(r_{ij_m}=1|\mathbf{x}_i,y_i,\boldsymbol{\phi}_{j_m})$ can be modeled straightforwardly by a logistic regression model, such that $$\text{logit}[p(r_{ij_m}=1|\mathbf{x}_i,y_i,\boldsymbol{\phi}_{j_m})]=\phi_{0j_m}+y_i\phi_{1j_m}+\mathbf{x}_i^o\boldsymbol{\phi}_{2j_m}+\mathbf{x}_i^m\boldsymbol{\phi}_{3j_m},$$ where $\phi_{0j_m}$ is the intercept of the $j_m^{th}$ missingness model, $\phi_{1j_m}$ is the coefficient pertaining to the response variable $\mathbf{Y}$, and $\boldsymbol{\phi}_{2j_m}=\{\phi_{2,j_m,1},\ldots,\phi_{2,j_m,p_{obs}^X}\}^T$ and $\boldsymbol{\phi}_{3j_m}=\{\phi_{3,j_m,1},\ldots,\phi_{3,j_m,p_{miss}^X}\}^T$ are the sets of coefficients of the $j_m^{th}$ variable's missingness model pertaining to the effects of the observed and missing features on the missingness, respectively, with $p_{obs}^X$ and $p_{miss}^X$ denoting the number of completely-observed and partially observed features in $\mathbf{X}$, respectively. Note that this model assumes independence of $\mathbf{R}$ across the $p_{miss}$ missing features, such that the missingness of each variable is conditionally independent of whether any other variable has been observed, which may or may not be realistic in some settings \citep{Ibrahim2005}.

When missingness is assumed to be MAR or MCAR, the marginal log-likelihood can be factored as $\log p_{\alpha, \boldsymbol{\beta}, \pi, \psi, \phi}(\mathbf{X}^o,\mathbf{Y}^o,\mathbf{R}) = \log p_{\alpha,\boldsymbol{\beta}, \pi,\psi}(\mathbf{X}^o,\mathbf{Y}^o) + \log p_\phi(\mathbf{R}|\mathbf{X}^o,\mathbf{Y^o})$. In this case, the quantity $\log p_\phi(\mathbf{R}|\mathbf{X}^o,\mathbf{Y^o})$ need not be specified, since it is independent from the parameters of interest pertaining to $p_{\alpha,\boldsymbol{\beta}, \pi,\psi}(\mathbf{X}^o,\mathbf{Y}^o)$.  Therefore, MAR or MCAR missingness is often referred to as ``ignorable'' missingness. Equation (\ref{eqn:LLNonignorable3}) can then be expressed as
\begin{equation}
\log p_{\alpha,\boldsymbol{\beta}, \pi,\psi}(\mathbf{X}^o,\mathbf{Y}^o)=\log \iint p_{\alpha,\boldsymbol{\beta}, \pi}(\mathbf{Y}|\mathbf{X}) p_{\psi}(\mathbf{X}) d\mathbf{X}^m d\mathbf{Y}^m.
\label{eqn:LLignorable}
\end{equation}

\subsection{Deeply-learned GLM with Missingness (\textit{dlglm})} \label{sec:dlglm}
In this section, we propose an algorithm for training deeply-learned GLMs in the presence of MCAR, MAR, and MNAR missingness. Before discussing this model, we first discuss the specification of the so-called covariate distribution $p_\psi(\mathbf{X})$ introduced in Equations \ref{eqn:LLNonignorable3} and \ref{eqn:LLignorable}, which is critical for maximizing the marginal log-likelihood in either setting. In Sections \ref{sec:MVNcovar}-\ref{sec:vaeiwae}, we discuss two different models for $p_\psi(\mathbf{X})$, and then in Section \ref{sec:dlglm2} we propose a novel method to handle missingness using a deeply-learned GLM architecture with an Importance-Weighted Autoencoder (IWAE) covariate structure. To simplify the discussion, we narrow the scope of our discussion to the \textit{Case x} setting, where only $\mathbf{X}$ contains missingness, but note that the proposed methodology naturally extends to \textit{Case y} and \textit{Case xy} settings as well.  


\subsubsection{Modeling $p_\psi(\mathbf{X})$ with known distribution} \label{sec:MVNcovar}
Given Eq. \ref{eqn:LLNonignorable3}, we must model \textbf{X} with some assumed covariate distribution $p_\psi(\mathbf{X})$. Care must be taken in specifying this distribution, as improper specification may reduce the accuracy of estimation of the parameters of interest $\boldsymbol{\beta}$ \citep{Lipsitz1996}. For example, we may assume $p_\psi(\mathbf{X})$ follows some known multivariate distribution such as the multivariate normal distribution, where $\mathbf{X} \sim N_p(\boldsymbol{\mu},\boldsymbol{\Sigma})$ and $\psi = \{\boldsymbol{\mu},\boldsymbol{\Sigma}\}$. Here, $\psi$ can be optimized jointly with the rest of the parameters $\{\alpha, \boldsymbol{\beta}, \pi, \phi\}$ that are involved in the marginal log-likelihood. However, this assumption may not be applicable in many instances such as in the case when $\mathbf{X}$ contains mixed data types, where both continuous and discrete features may be correlated and a joint distribution may be difficult to specify in closed form.  In certain cases, it may be beneficial to model $p_\psi(\mathbf{X})$ flexibly, such that no strong prior assumptions need to be made on the form of this distribution. To address this, a sequence of 1-D conditionals have previously been proposed to model the covariate distribution \citep{Lipsitz1996}, but such a model may be computationally intractable when the number of covariates is very large.

Once an explicit form for the covariate distribution is specified, one aims to maximize the marginal log-likelihood, as introduced in Equation (\ref{eqn:LLNonignorable3}) in Section \ref{sec:missglm}. However, due to the integral involved, this quantity is often intractable and is difficult to maximize directly, so a lower bound of the marginal log-likelihood is often maximized instead. The derivation of this lower bound can be found in Appendix A2 of the supplementary materials.  


\subsubsection{Modelling $p_\psi(\mathbf{X})$ with Variational and Importance-Weighted Autoencoders} \label{sec:vaeiwae}

Alternatively, one can approximately learn $p_\psi(\mathbf{X})$ from the training data by using an IWAE neural network architecture. In this section, we first introduce a general form of the variational autoencoder (VAE) and IWAE in the case of completely-observed data $\mathbf{X}$. Then, in Section  \ref{sec:dlglm2}, we apply the IWAE covariate structure to the deeply-learned GLM setting and show how this representation naturally extends to the case where MCAR, MAR, or MNAR missingness is observed in $\mathbf{X}$ when training deeply-learned GLMs.

First, let $\mathbf{Z}$ be an $n \times d$ matrix, such that $\mathbf{Z}=\{\mathbf{z}_1,\cdots,\mathbf{z}_n\}$ and $\mathbf{z}_i$ is a latent vector of length $d$ pertaining to the $i^{th}$ sample latent variable, and let $\mathbf{Z}$ represent a lower-dimensional representation or subspace of $\mathbf{X}$. It is common practice to tune the value of $d$ as a hyperparameter by choosing the optimal integer value that best fits the data, as measured by some objective function. In a VAE, we assume $\mathbf{x}_1,\ldots,\mathbf{x}_n$ are i.i.d. samples from a multivariate p.d.f or ``generative model" $p_\psi(\mathbf{X} | \mathbf{Z})$ with accompanying parameters $\psi$ that describes how $\mathbf{X}$ is generated from the lower dimensional space $\mathbf{Z}$. In this manner, a VAE aims to learn accurate representations of high-dimensional data, and may be used to generate synthetic data with similar qualities as the training data. These aspects are also aided through the use of embedded deep learning neural networks, for example within $p_\psi(\mathbf{X} | \mathbf{Z})$, which also facilitates its applicability to larger dimensions and complex datasets.


In a VAE with completely observed training data, one aims to maximize the marginal log-likelihood as $\log p_\psi(\mathbf{X})=\log \int p_\psi(\mathbf{X},\mathbf{Z}) d\mathbf{Z} = \log \int p_\psi(\mathbf{X} | \mathbf{Z})p(\mathbf{Z}) d\mathbf{Z}$. However, this quantity is also often intractable and difficult to maximize directly. Therefore, VAE's alternatively optimize an objective function called the ``Evidence Lower Bound" (ELBO), which lower bounds $\log p_\psi(\mathbf{X})$ and has the following form \citep{Kingma2013}:
\begin{align*}
\mathcal{L}^{ELBO}(\theta,\psi) &= \E_{\mathbf{Z} \sim q_\theta(\mathbf{Z}|\mathbf{X})} \log \left[ \frac{p_\psi(\mathbf{X}|\mathbf{Z})p(\mathbf{Z})}{q_\theta(\mathbf{Z}|\mathbf{X})} \right]\stepcounter{equation}\tag{\theequation}\label{eqn:ELBO1} \\
\hat{\mathcal{L}}^{ELBO}_K(\theta,\psi) &= \frac{1}{K} \sum_{k=1}^K \log \left[ \frac{p_\psi(\mathbf{X}|\tilde{\mathbf{Z}}_k)p(\tilde{\mathbf{Z}}_k)}{q_\theta(\tilde{\mathbf{Z}}_k|\mathbf{X})} \right].\stepcounter{equation}\tag{\theequation}\label{eqn:ELBO2}
\end{align*}

Here, $\mathcal{L}^{ELBO}(\theta,\psi)$ denotes the ELBO such that $\mathcal{L}^{ELBO}(\theta,\psi) \leq \log p_\psi(\mathbf{X})$.   Also let $\hat{\mathcal{L}}^{ELBO}_K(\theta,\psi)$ denote the empirical approximation to Eq. (\ref{eqn:ELBO1}) computed by Monte Carlo integration, such that  $\mathcal{L}^{ELBO}(\theta,\psi) \approx \hat{\mathcal{L}}_K^{ELBO}(\theta,\psi)$ and  $\tilde{\mathbf{Z}}_1,\ldots,\tilde{\mathbf{Z}}_K$ are $K$ samples drawn from $q_\theta(\mathbf{Z}|\mathbf{X})$, the variational approximation of the true but intractable posterior $p_\psi(\mathbf{Z}|\mathbf{X})$, also called the ``recognition model". Furthermore,  denote $f_\psi(\mathbf{Z})$ and $g_\theta(\mathbf{X})$ as the decoder and encoder feed forward neural networks of the VAE, where $\psi$ and $
\theta$ are the sets of weights and biases pertaining to each of these neural networks, respectively. Given $\mathbf{Z}$, $f_\psi(\mathbf{Z})$ outputs the distributional parameters pertaining to $p_\psi(\mathbf{X}|\mathbf{Z})$.  

In variational inference, $q_\theta(\mathbf{Z}|\mathbf{X})$ is constrained to be from a class of simple distributions, or ``variational family", to obtain the best candidate from within that class to approximate $p_\psi(\mathbf{Z}|\mathbf{X})$. Variational inference is usually used in tandem with amortization  of the parameters where the neural network parameters are shared across observations \citep{Gershman2014}, allowing for stochastic gradient descent (SGD) to be used for optimization of Eq. (\ref{eqn:ELBO2}) \citep{Kingma2019}. In practice, both $q_\theta(\mathbf{Z}|\mathbf{X})$ and $p(\mathbf{Z})$ are typically assumed to have simple forms, such as multivariate Gaussians with diagonal covariance structures, and $q_\theta(\mathbf{Z}|\mathbf{X})$ is commonly assumed to be factorizable, such that $q_\theta(\mathbf{Z}|\mathbf{X}) = \prod_{i=1}^n q_\theta(\mathbf{z}_i|\mathbf{x}_i)$ \citep{Kingma2019}. Although one can specify a class of more complicated distributions for $q_\theta(\mathbf{Z}|\mathbf{X})$ as long as they are reparameterizable \citep{Li2020,Strauss2021,Strauss2022}, the multivariate Gaussian with diagonal covariance structure is most often used, following works by \citet{Burda2015} and \citet{Kingma2013}, due to the convenience in sampling and computation \citep{Kingma2019}.


Let $(\hat{\theta}^{(t)},\hat{\psi}^{(t)})$ be the estimates of $(\theta,\psi)$ at update (or iteration) $t$. For $t=0$, these values are often initialized to small values centered around 0, although other initialization schemes may be used \citep{Saxe2014,Murphy2016}.  Each subsequent update $t\geq 1$ consists of two general steps to maximize $\mathcal{L}(\theta,\psi)$. First, $K$ samples are drawn from $q_{\hat{\theta}^{(t)}}(\mathbf{Z}|\mathbf{X})$ to compute the quantity in Eq. (\ref{eqn:ELBO2}), conditional on $\hat{\theta}^{(t)}$, similar to importance sampling. Then, the so-called ``reparametrization trick" is utilized to facilitate the calculation of gradients of this approximation to obtain $(\hat{\theta}^{t+1},\hat{\psi}^{t+1})$ using stochastic gradient descent \citep{Kingma2013}. The networks $f_\psi(\mathbf{Z})$ and $g_\theta(\mathbf{X})$ also allow the VAE to capture complex and non-linear relationships between features in outputting the distributional parameters for the generative and recognition models, respectively. This procedure may be repeated for a fixed number of iterations, or may be terminated early due to pre-specified convergence criteria \citep{Prechelt1998}. \citet{Kingma2013} provides additional details on the maximization procedure for VAEs.  


The IWAE \citep{Burda2015} is a generalization of the standard VAE. Both the VAE and IWAE estimate $\log p_\psi(\mathbf{X})$ by drawing samples of latent variables to estimate an expectation. However, while the VAE utilizes $p_\psi(\mathbf{X},\mathbf{Z})/q_\theta(\mathbf{Z}|\mathbf{X})$ as the importance weights in deriving the ELBO, the IWAE uses the average of $K$ importance weights in the integrand for a tighter lower bound of the marginal log-likelihood \citep{Burda2015}. The resulting IWAE bound, corresponding to the ELBO in Eq. (\ref{eqn:ELBO1}), can be written as
\begin{align*}
\mathcal{L}_K^{IWAE}(\theta,\psi)&= \E_{\mathbf{Z}_k \sim q_\theta(\mathbf{Z}|\mathbf{X})} \log \left[ \frac{1}{K} \sum_{k=1}^K \frac{p_\psi(\mathbf{X}|\mathbf{Z}_k)p(\mathbf{Z}_k)}{q_\theta(\mathbf{Z}_k|\mathbf{X})} \right] \stepcounter{equation}\tag{\theequation}\label{eqn:IWAEbound1}\\
\hat{\mathcal{L}}_K^{IWAE}(\theta,\psi)&= \log \left[ \frac{1}{K} \sum_{k=1}^K \frac{p_\psi(\mathbf{X}|\tilde{\mathbf{Z}}_k)p(\tilde{\mathbf{Z}}_k)}{q_\theta(\tilde{\mathbf{Z}}_k|\mathbf{X})} \right].\stepcounter{equation}\tag{\theequation}\label{eqn:IWAEbound2}
\end{align*}
Importantly, although $K$ samples are drawn from $q(\mathbf{Z}|\mathbf{X})$ to estimate the lower bound for both the VAE and IWAE, a VAE assumes a single latent variable $\mathbf{Z}$ that is sampled $K$ times, wheras an IWAE assumes $\mathbf{Z}_1,\ldots,\mathbf{Z}_K$ are independent and identically distributed (i.i.d.) latent variables, and each variable is sampled once from $q(\mathbf{Z}|\mathbf{X})$. Typically, just one sample is drawn for each latent variable to estimate the ELBO and IWAE bound. If $K=1$, $\mathcal{L}_1^{IWAE}=\mathcal{L}^{VAE}$, and the IWAE corresponds exactly to the standard VAE. For $K>1$, \citet{Burda2015} showed that $\log p(\mathbf{X}) \geq \hat{\mathcal{L}}_{K+1}^{IWAE} \geq \hat{\mathcal{L}}_K^{IWAE}$, such that $\hat{\mathcal{L}}_K^{IWAE} \rightarrow \log p(\mathbf{X})$ as $K \rightarrow \infty$ if $p_\psi(\mathbf{X},\mathbf{Z})/q_\theta(\mathbf{Z}|\mathbf{X})$ is bounded. Thus, the IWAE bound more closely approximates the true marginal log likelihood when $K>1$ \citep{Cremer2017}, but the computational burden is increased due to the increased number of samples. A visualization of the workflow for an IWAE can be found in Appendix A3 of the supplementary materials. 

\subsubsection{dlglm: Modeling X in the presence of missingness}\label{sec:dlglm2}

Now, we extend the above framework to the deeply-learned GLM framework, where features within $\mathbf{X}$ are partially observed during training. We formally introduce the \textit{dlglm} model to handle MNAR missingness in the context of deeply-learned GLMs, as well as a variant of \textit{dlglm} to specifically handle MCAR and MAR missingness. 


Let us define $q_\theta(\mathbf{Z},\mathbf{X}^m)$ as the variational joint posterior pertaining to $(\mathbf{Z},\mathbf{X}^m)$. Then, we can factor this variational joint posterior as $q_\theta(\mathbf{Z},\mathbf{X}^m) = q_{\theta_1}(\mathbf{Z}|\mathbf{X}^o)q_{\theta_2}(\mathbf{X}^m|\mathbf{Z},\mathbf{X}^o,\mathbf{R},\mathbf{Y})$. Here, for $k=1,\ldots,K$, we assume $\mathbf{Z}_k \stackrel{i.i.d}{\sim} q_{\theta_1}(\mathbf{Z}|\mathbf{X}^o)$ similar to an IWAE, and additionally assume $\mathbf{X}_{k}^m \stackrel{i.i.d}{\sim} q_{\theta_2}(\mathbf{X}^m|\mathbf{Z},\mathbf{X}^o,\mathbf{R},\mathbf{Y})$, where each $\mathbf{X}_{k}^m$ has dimensionality $p_{miss}^X$. Here, we assume that \textbf{Y} is generated by \textbf{X}, and thus it is redundant to utilize \textbf{Y} in the part of the variational joint posterior pertaining to \textbf{Z}. Empirically, we observed that including \textbf{Y} in the conditional, such that $q(\mathbf{Z}|\mathbf{X}^o,\mathbf{Y})$, did not have a significant impact. Additionally, we note that the form $q_{\theta_2}(\mathbf{X}^m|\mathbf{Z},\mathbf{X}^o,\mathbf{R},\mathbf{Y})$ includes $\mathbf{Y}$, allowing for more accurate imputation of missing values; however, we remove this term in the conditional in the context of prediction, in order to predict $\mathbf{Y}$ in an unbiased manner.

We then utilize the class of factored variational posteriors, such that $q_{\theta}(\mathbf{Z},\mathbf{X}^m)=\prod_{i=1}^{n}q_{\theta}(\mathbf{z}_i,\mathbf{x}_i^m)$ and  $q_{\theta}(\mathbf{z}_i,\mathbf{x}_i^m)=q_{\theta_1}(\mathbf{z}_i|\mathbf{x}_i^o)q_{\theta_2}(\mathbf{x}_i^m|\mathbf{z}_i,\mathbf{x}_i^o,\mathbf{r}_i^X)$, with $\theta = \{\theta_1,\theta_2\}$. Then, denoting $\mathbf{z}_{ik}$ and $\mathbf{x}_{ik}^m$ as the $i^{th}$ observation vectors of $\mathbf{Z}_k$ and $\mathbf{X}_k^m$, respectively, we have $\mathbf{z}_{i1},\ldots,\mathbf{z}_{iK} \stackrel{i.i.d}{\sim} q_{\theta_1}(\mathbf{z}_i|\mathbf{x}_i^o)$ and $\mathbf{x}_{i1}^m,\ldots,\mathbf{x}_{iK}^m \stackrel{i.i.d}{\sim} q_{\theta_2}(\mathbf{x}_i^m|\mathbf{z}_i,\mathbf{x}_i^o,\mathbf{r}_i^X)$. In this case, the lower bound, which we call the ``dlglm bound'', can be derived as follows:
\begin{align*}
\log p&_{\alpha, \boldsymbol{\beta}, \pi, \psi, \phi}(\mathbf{X}^o,\mathbf{Y},\mathbf{R}^X) = \sum_{i=1}^{n} \log p_{\alpha, \boldsymbol{\beta}, \pi, \psi, \phi}(\mathbf{x}_i^o,y_i,\mathbf{r}_i^X) \\
&= \sum_{i=1}^n \log \left[\iint p_{\alpha, \boldsymbol{\beta}, \pi, \psi, \phi}(\mathbf{x}_i^o,\mathbf{x}_i^m,y_i,\mathbf{r}_i^X,\mathbf{z}_i) d\mathbf{z}_i d\mathbf{x}_i^m \right]\\
&= \sum_{i=1}^n \log \E_{(\mathbf{z}_{ik},\mathbf{x}_{ik}^m) \sim q_{\theta}(\mathbf{z}_i,\mathbf{x}_i^m)} \left[ \frac{1}{K}\sum_{k=1}^K  \frac{p_{\alpha, \boldsymbol{\beta}, \pi, \psi, \phi}(\mathbf{x}_i^o,\mathbf{x}_{ik}^m,y_i,\mathbf{r}_i^X,\mathbf{z}_{ik})}{q_\theta(\mathbf{z}_{ik},\mathbf{x}_{ik}^m)} \right]\\
&\geq \sum_{i=1}^{n} \E_{(\mathbf{z}_{ik},\mathbf{x}_{ik}^m) \sim q_{\theta}(\mathbf{z}_i,\mathbf{x}_i^m)} \log{ \left[ \frac{1}{K}\sum_{k=1}^{K}\frac{p_{\alpha, \boldsymbol{\beta}, \pi, \psi, \phi}(\mathbf{x}_i^o,\mathbf{x}_{ik}^m,y_i,\mathbf{r}_i^X,\mathbf{z}_{ik})}{q_{\theta}(\mathbf{z}_{ik},\mathbf{x}_{ik}^m)} \right] } = \mathcal{L}_{K}^{dlglm},
\stepcounter{equation}\tag{\theequation}\label{eqn:dlglmbound1}
\end{align*}

Here, $\{\psi,\boldsymbol{\beta}, \pi,\phi,\theta\}$ are the weights and biases associated with the neural networks that output the parameters of the distributions that are involved, $\alpha$ is the dispersion parameter associated with the variance function of $\mathbf{Y}$, and $\tilde{\mathbf{z}}_{ik}$ and $\tilde{\mathbf{x}}_{ik}^m$ are the samples drawn from $q_{\theta_1}(\mathbf{z}_i|\mathbf{x}_i^o)$, and $q_{\theta_2}(\mathbf{x}_i^m|\mathbf{z}_i,\mathbf{x}_i^o,\mathbf{r}_i^X)$, respectively.

As discussed in Section \ref{sec:missglm}, we use the selection model factorization of the complete data log-likelihood, such that $p_{\alpha, \boldsymbol{\beta}, \pi, \psi, \phi}(\mathbf{x}_i^o,\mathbf{x}_i^m, y_i, \mathbf{r}_i^X, \mathbf{z}_i)=p_{\alpha,\boldsymbol{\beta}, \pi}(y_i|\mathbf{x}_i)p_\psi(\mathbf{x}_i|\mathbf{z}_i)p(\mathbf{z}_i)p_\phi(\mathbf{r}_i^X|\mathbf{x}_i,y_i).$ As before, we can remove $y_i$ from $p_\phi(\mathbf{r}_i^X|\mathbf{x}_i,y_i)$ for unbiased prediction. Then, applying this factorization to (\ref{eqn:dlglmbound1}), we obtain the form of the estimate of the ``dlglm bound'', where the integral is estimated via Monte Carlo integration:
\begin{equation}
\hat{\mathcal{L}}_{K}^{dlglm} = \sum_{i=1}^{n}  \log{\left[\frac{1}{K} \sum_{k=1}^{K} \frac{p_{\alpha,\boldsymbol{\beta}, \pi}(y_i|\mathbf{x}_i^o,\tilde{\mathbf{x}}_{ik}^m)p_{\psi}(\mathbf{x}_i|\tilde{\mathbf{z}}_{ik})p(\tilde{\mathbf{z}}_{ik})p_{\phi}(\mathbf{r}_i^X|\mathbf{x}_i^o,\tilde{\mathbf{x}}_{ik}^m)}{q_{\theta_1}(\tilde{\mathbf{z}}_{ik}|\mathbf{x}_i^o) q_{\theta_2}(\tilde{\mathbf{x}}_{ik}^m|\tilde{\mathbf{z}}_{ik},\mathbf{x}_{i}^o,\mathbf{r}_i^X)} \right] } , 
\label{eqn:dlglmbound2}
\end{equation}

\begin{figure}[H]
\begin{center}
\includegraphics[width=165mm]{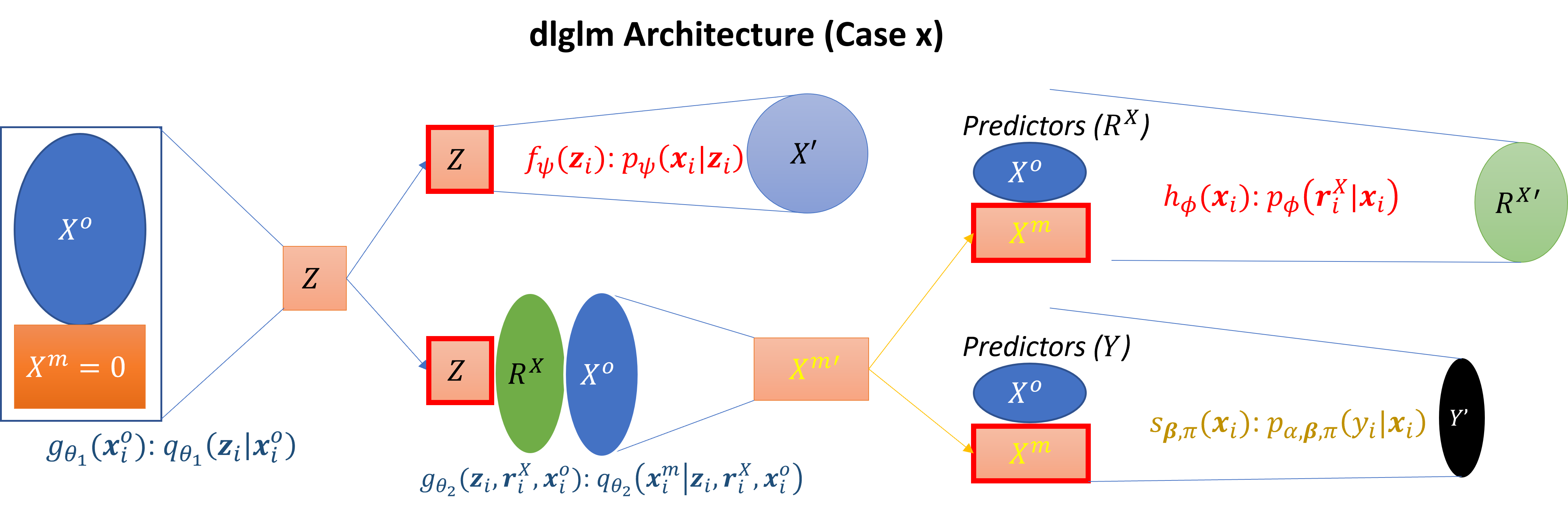} 
\end{center}
\caption{\scriptsize Architecture of proposed dlglm method (\textit{Case x}). Dark colored nodes ($X^o, X^m=0, R^X$) represent deterministic values, lightly colored nodes ($Z^{\prime}, X^{o\prime},X^{m\prime}, R^{X\prime},Y^{\prime}$) represent learned distributional parameters, and outlined (in red) nodes represent sampled values. Orange cells correspond to latent variables $\mathbf{Z}$ and $\mathbf{X}^m$. $\mathbf{Z}_1,\ldots, \mathbf{Z}_K$ and $\mathbf{X}_{1}^m,\ldots,\mathbf{X}_{K}^m$ are sampled from their respective variational posteriors $q_{\theta_1}(\mathbf{Z}|\mathbf{X}^o)$ and $q_{\theta_2}(\mathbf{X}^m|\mathbf{Z},\mathbf{R}^X,\mathbf{X}^o)$.}
\label{fig:dlglmarchitecture}
\end{figure}

We see that this quantity closely resembles the lower bound of an IWAE, and, similar to traditional VAEs, we utilize neural networks $f_{\psi}(\mathbf{z}_i)$, $g_{\theta_1}(\mathbf{x}_i^o)$, $g_{\theta_2}(\mathbf{z}_i,\mathbf{x}_i^o,\mathbf{r}_i^X)$, $s_{\boldsymbol{\beta}, \pi}(\mathbf{x}_i)$, and $h_{\phi}(\mathbf{x}_i)$ to learn the values of the parameters of $p_\psi(\mathbf{x}_i|\mathbf{z}_i)$, $q_{\theta_1}(\mathbf{z}_i|\mathbf{x}_i^o)$, $q_{\theta_2}(\mathbf{x}_i^m|\mathbf{z}_i,\mathbf{x}_i^o,\mathbf{r}_i^X)$, $p_{\alpha,\boldsymbol{\beta}, \pi}(y_i|\mathbf{x}_i)$, and $p_\phi(\mathbf{r}_i^X|\mathbf{x}_i)$. The associated weights and biases of the neural networks $\{\boldsymbol{\beta}, \pi, \psi, \phi\}$, as well as the dispersion parameter $\alpha$ pertaining to $p_{\alpha,\boldsymbol{\beta}, \pi}(\mathbf{Y}|\mathbf{X})$ are updated using stochastic gradient descent via the ADAM optimizer \citep{Kingma2014}. Importantly, we call the neural network denoted by $h_\phi(\mathbf{x}_i)$ the ``missingness network". The inclusion of this network allows us to learn a model for the missingness mechanism, which is essential for accurate analysis in the presence of MNAR or non-ignorable missingness. The architecture of \textit{dlglm} can be found in Figure \ref{fig:dlglmarchitecture}. A pseudo-algorithm of \textit{dlglm} can be found in Appendix A4 of the supplementary materials. We limited our discussion in this paper to \textit{Case x}, where missingness exists only in $\mathbf{X}$ but not in \textbf{Y}; however, the lower bound for \textit{dlglm} can similarly be derived for the more general \textit{Case xy} as well, and this derivation can be found in Appendix A5 of the supplementary materials.

We can obtain a variant of this method, which we call ignorably-missing dlglm (\textit{idlglm}), by assuming independence between $\mathbf{X}^m$ and $\mathbf{R}$ by omitting $\mathbf{r}_i^X$ from Equation \ref{eqn:dlglmbound1}, and removing $p_\phi(\mathbf{r}_i^X|\mathbf{x}_i^o,\tilde{\mathbf{x}}_{ik}^m)$ and letting $p_\phi(\tilde{\mathbf{x}}_{ik}^m|\tilde{\mathbf{z}}_{ik},\mathbf{x}_i^o,\mathbf{r}_i^X) \rightarrow p_\phi(\tilde{\mathbf{x}}_{ik}^m|\tilde{\mathbf{z}}_{ik},\mathbf{x}_i^o)$ in Equation \ref{eqn:dlglmbound2}. Whereas \textit{dlglm} is better suited to handle MNAR, \textit{idlglm} may be more appropriate for the MCAR or MAR settings, where a missingness model need not be specified.

In this paper, we are primarily interested in supervised learning. However, following training, \textit{dlglm} and \textit{idlglm} can also perform imputation as in the unsupervised learning architecture for handling missingness proposed by \citet{Lim2021}, although such imputation is not necessary for training, coefficient estimation, or prediction. The single imputation procedure, and additional computational details of \textit{dlglm} and \textit{idlglm} can be found in Appendix A6 and A1 of the supplementary materials, respectively.

A recently published method by \citet{Ipsen2021} performs unsupervised learning by similarly learning a missingness model in their neural network framework to handle MNAR missingness. However, they assume that $p(\mathbf{X}^m|\mathbf{Z}) \approx q(\mathbf{X}^m|\mathbf{Z},\mathbf{X}^o,\mathbf{R})$. This may be an oversimplification, as in the MNAR case, \textbf{R} cannot be assumed to be independent of $\mathbf{X}^m$. Recent work by \citet{Ma2021} similarly performs unsupervised learning under MNAR missingness, including an auxiliary fully-observed variable to guarantee identifiability. However, they also make the same simplifying assumption as \citet{Ipsen2021}, which may not hold in MNAR. In addition, both methods are designed for imputation, rather than supervised tasks, and extending these methods for such tasks may be nontrivial, especially for computationally intensive models. Issues of identifiability in missing data applications often lead to issues of convergence during model training \citep{Beesley2019}. We note that although deriving the identifiability of \textit{dlglm} is not focal point of this paper, we consistently observed convergence in training the \textit{dlglm} architecture in various simulations and real data settings.

\section{Numerical Examples} \label{sec:num3}
In this section, we evaluate the performance of \textit{dlglm} and \textit{idlglm} to analyze each method's performance in imputation, coefficient estimation, and prediction tasks on simulated datasets under MCAR, MAR, and MNAR missingness in Section \ref{sec:sim3}. We also compare our methods to two commonly used approaches for modeling missing data in the supervised setting, mean imputation and the \textit{mice} method for multiple imputation \citep{VanBuuren2011}. We also compared performance in simulated data with two deep learning methods that were recently published \textit{miwae} \citep{Mattei2019} and \textit{notmiwae} \citep{Ipsen2021}. To account for potential non-linearity and complex relationships between features, in Section \ref{sec:real3}, we mask completely-observed datasets obtained from the UCI Machine Learning Repository with varying mechanisms of missingness on the predictors.  Finally, in Section \ref{sec:bankmarketing}, we perform prediction on the Bank Marketing dataset, which inherently contains missingness in the predictors.

In all simulated and real data analyses, we tuned a variety of hyperparameters for deep learning methods, including the number of hidden layers, the dimensionality of the latent variable $\mathbf{Z}$, and the number of nodes per hidden layer. For \textit{dlglm}, we additionally tuned the number of hidden layers in the missingness network separately, allowing the network to accurately capture potentially complex nonlinear relationships in the missingness model.

A grid-search approach was used for training based upon discrete pre-specified values, selecting the optimal combination of hyperparameters using the lower bound computed on a held out validation set. The selected hyperparameters for the simulated datasets, as well as the UCI and Bank Marketing datasets are listed in Appendix B1 of the supplementary materials.

\subsection{Simulated Data} \label{sec:sim3}


\subsubsection{Simulation Setup} \label{sec:simsetup3}
We first utilized completely synthetic data to evaluate the performance of each. Here, $\mathbf{X}$ is generated such that $\mathbf{X} = normalize(\mathbf{Z}\mathbf{W} + \mathbf{B}) + B_0$, where $normalize(\cdot)$ takes an input matrix and standardizes each column to mean 0 and standard deviation 1, and $\mathbf{W}$ and $\mathbf{B}$ and are matrices of dimensions $d \times p$ and $n \times p$, respectively, $\mathbf{Z}\sim N_{d}(\mathbf{0},\mathbf{I})$, and $W_{lj} \sim N(0,0.5)$ and $B_{ij} \sim N(0,1)$ for $i = 1,\ldots,n$, $p = 1,\ldots,p$, and $l = 1,\ldots,d$, and $B_0=2$ is fixed. We also generated a binary response variable $\mathbf{Y}$ such that logit$[$Pr$(\mathbf{Y}=1|\mathbf{X})] = \beta_0 + \boldsymbol{\beta}\mathbf{X}$, where $\boldsymbol{\beta}$ are drawn randomly from $\{-\frac{1}{4},\frac{1}{4}\}$, and $\beta_0$ is chosen such that approximately half of the sample are in either class. Values of $\mathbf{Y}$ are drawn from Bernoulli(Pr$(\mathbf{Y}=1|\mathbf{X})$).

We then simulate the missingness mask matrix $\mathbf{R}^X$ such that 50\% of features in $\mathbf{X}$ are partially observed, and 30\% of the observations for each of these features are missing. We generate $r_{ij}$ from the Bernoulli distribution with probability equal to $p(r_{ij_m}=1|\mathbf{x}_i,y_i,\boldsymbol{\phi})$, such that $\text{logit} [p(r_{ij_m}=1|\mathbf{x}_i,y_i,\boldsymbol{\phi})]=\phi_0+\phi_1 y_i+\boldsymbol{\phi}_2\mathbf{x}_i^o + \boldsymbol{\phi}_3\mathbf{x}_i^m$, where $j_m=1,\ldots,p_{miss}^X$ index the missing features, $\phi_1$ is the coefficient pertaining to the response, $\boldsymbol{\phi}_2 = \{\phi_{21},\ldots,\phi_{2,p_{obs}^X}\}$ are the coefficients pertaining to the observed features, and $\boldsymbol{\phi}_3 = \{\phi_{31},\ldots,\phi_{3,p_{miss}^X}\}$ are those pertaining to the missing features, where $p_{obs}^X$ and $p_{miss}^X$ are the total number of features that are observed and missing, respectively, with $p_{miss}^X = floor(0.5*p)$ and $p_{obs}^X=p-p_{miss}^X$. Here, we fixed $\phi_1=0$, and drew nonzero values of $\{\boldsymbol{\phi}_2,\boldsymbol{\phi}_3\}$ from the log-normal distribution with mean $\mu_\phi=5$, with $\log$ standard deviation $\sigma_\phi=0.2$.  

To evaluate the impact of the misspecification of the missingness mechanism on model performance, $r_{ij_m}$ was simulated under each mechanism as follows: (1) MCAR: $\{\phi_1,\boldsymbol{\phi}_2,\boldsymbol{\phi}_3\}=0$ (2) MAR: Same as MCAR except $\phi_{2j_o} \neq 0$ for one completely-observed feature $j_o$ (3) MNAR: Same as MCAR except $\phi_{3j_m} \neq 0$ for one missing feature $j_m$. In this way, for each MAR or MNAR feature, the missingness is dependent on just one feature. In each case, we used $\phi_0$ to control for an expected rate of missingness of $30\%$ in each partially-observed feature. We note that for each these simulations, we utilize all features in $\mathbf{X}$ as well as the response $\mathbf{Y}$ as input into \textit{dlglm}'s missingness network, although only one feature is involved under the true missingness model. Additionally, we searched for the optimal variational distributions of $q_{\theta_1}(\mathbf{Z}|\mathbf{X}^o)$ and $q_{\theta_2}(\mathbf{X}^m|\mathbf{Z},\mathbf{X}^o,\mathbf{R})$ from a class of Gaussian distributions with diagonal covariance structures, as discussed in Section \ref{sec:vaeiwae}. We fixed $K=5$ during training, and increased $K$ to $500$ at test time.

We vary $n$ and $d$ such that $n=\{10,000, 100,000\}$ and $d=\{2,8\}$, and fix $p=50$. We simulated 5 datasets per simulation condition, spanning various missingness mechanisms and values for $\{n,d\}$. We fix the values of $\boldsymbol{\beta}$ at $0.25$ for each feature, and adjusted $\beta_0$ to ensure equal proportions for the binary class response $\mathbf{Y}$. For each simulation case, we partitioned the data into training, validation, and test sets with ratio 8:1:1. For \textit{mice} imputation, we averaged across 500 multiply-imputed datasets to obtain a single imputed dataset. We note that we generated $\mathbf{Y}$ by a linear transformation of $\mathbf{X}$ in these simulations in order to facilitate fair comparisons with \textit{mice}, which cannot account for non-linear relationships between the features and the response. Because no hyperparameter tuning is required, the validation set is not utilized for \textit{mice} and mean imputation. 

We measured the performance of each method with respect to three different tasks: imputation of missing values, coefficient estimation, and prediction. Imputation performance was measured with respect to the truth on a single imputed dataset by mean, \textit{dlglm} and \textit{idlglm} imputation, and on an average of multiply-imputed datasets by \textit{mice}. Coefficient estimation for mean, \textit{miwae}, \textit{notmiwae}, and \textit{mice} were based on downstream fitted GLM(s) on these imputed dataset(s), where estimates were pooled using Rubin's rules \citep{Rubin2004} for \textit{mice}. For \textit{dlglm} and \textit{idlglm}, we estimated the coefficients by the weights and bias $\boldsymbol{\beta}$ of the last layer of the $s_{\boldsymbol{\beta},\pi}(\cdot)$ trained neural network.  Here, we fixed the number of hidden layers in $s_{\boldsymbol{\beta},\pi}(\cdot)$ to 0 to allow for direct comparison with the other methods. A more complex prediction model via a neural network can be learned by simply incorporating additional hidden layers in $s_{\boldsymbol{\beta},\pi}(\cdot)$. We note that  \textit{dlglm} and \textit{idlglm} can estimate $\boldsymbol{\beta}$ without having to perform multiple imputation and downstream modelling unlike \textit{mice}, where fitting complex methods such as neural networks each of the multiply-imputed datasets separately may be computationally prohibitive.

After obtaining the coefficient estimates and trained models, we performed prediction on the test set in two ways: 1) using the incomplete (predI) test set, where the true values of $\mathbf{X}^m$ are not known at prediction time, and 2) using the complete (predC) test set, where the true simulated values of $\mathbf{X}^m$ are known at prediction time. These two ways reflect the two realistic cases in which (1) missingness is present during training time but complete data is available at prediction time, and (2) missingness is present during both training and prediction time. For predI, \textit{miwae}, \textit{notmiwae}, \textit{mice} and mean imputation require an additional imputation step on the test set before predicting $\mathbf{Y}$; for \textit{dlglm} and \textit{idlglm}, we simply input the incomplete test set into the trained model without needing to separately impute the test set, and we predict using the trained model. That is, \textit{miwae}, \textit{notmiwae}, \textit{mice} and mean imputation cannot generalize the trained model to impute the test set, \textit{dlglm} and \textit{idlglm} provide a seamless framework to utilize the already-trained model to impute and predict on a held-out test set. For predC, we use the underlying true values of $\mathbf{X}^m$ to predict on the test dataset. 

Imputation error was measured by the average L1 distance between true and imputed masked values in $\mathbf{X}$. Letting $\hat{\mathbf{X}}^m$ denote the imputed masked values of the true $\mathbf{X}^m$ values of the missing entries, we denote the average L1 distance is simply  $\frac{\mid \hat{\mathbf{X}}^m - \mathbf{X}^m \mid}{N_{miss}},$  where $N_{miss}$ is the total number of missing entries in the dataset. Performance in coefficient estimation was measured by the average percent bias (PB) of the coefficient estimates compared to the truth, averaged across the $p$ features, i.e. $PB = 100 \times \frac{1}{p}\sum_{j=1}^p \frac{|\beta_j-\hat{\beta}_j|}{|\beta_j|}.$  Finally, predC and predI prediction error was measured by the average L1 distance between predicted and true values of the probabilities of class membership Pr$(\mathbf{Y}=1|\mathbf{X})$ in the test set.

In order to assess the sensitivity of the performance of these methods to the specification of the missingness model used to synthetically mask the data, we also repeated the analyses on data with missingness mask simulated by the following: $\text{logit} [p(r_{ij_m}=1|\mathbf{x}_i,y_i,\boldsymbol{\phi})]=\phi_0+\boldsymbol{\phi}_2\text{log}(\mathbf{x}_i^o + min(\mathbf{x}_i^o)) + \boldsymbol{\phi}_3\text{log}(\mathbf{x}_i^m + min(\mathbf{x}_i^m)),$ such that for the MAR and MNAR missingness cases, the missingness was dependent on the log of one of the completely or partially observed features. We denote this set of simulation conditions the ``nonlinear missingness" case, where the missingness was simulated from the specified nonlinear logistic regression model. We show the results of this analysis in Appendix B2 of the supplementary materials.

\subsubsection{Simulation Results} \label{sec:simres3}

\begin{figure}
\begin{center}
\includegraphics[width=155mm]{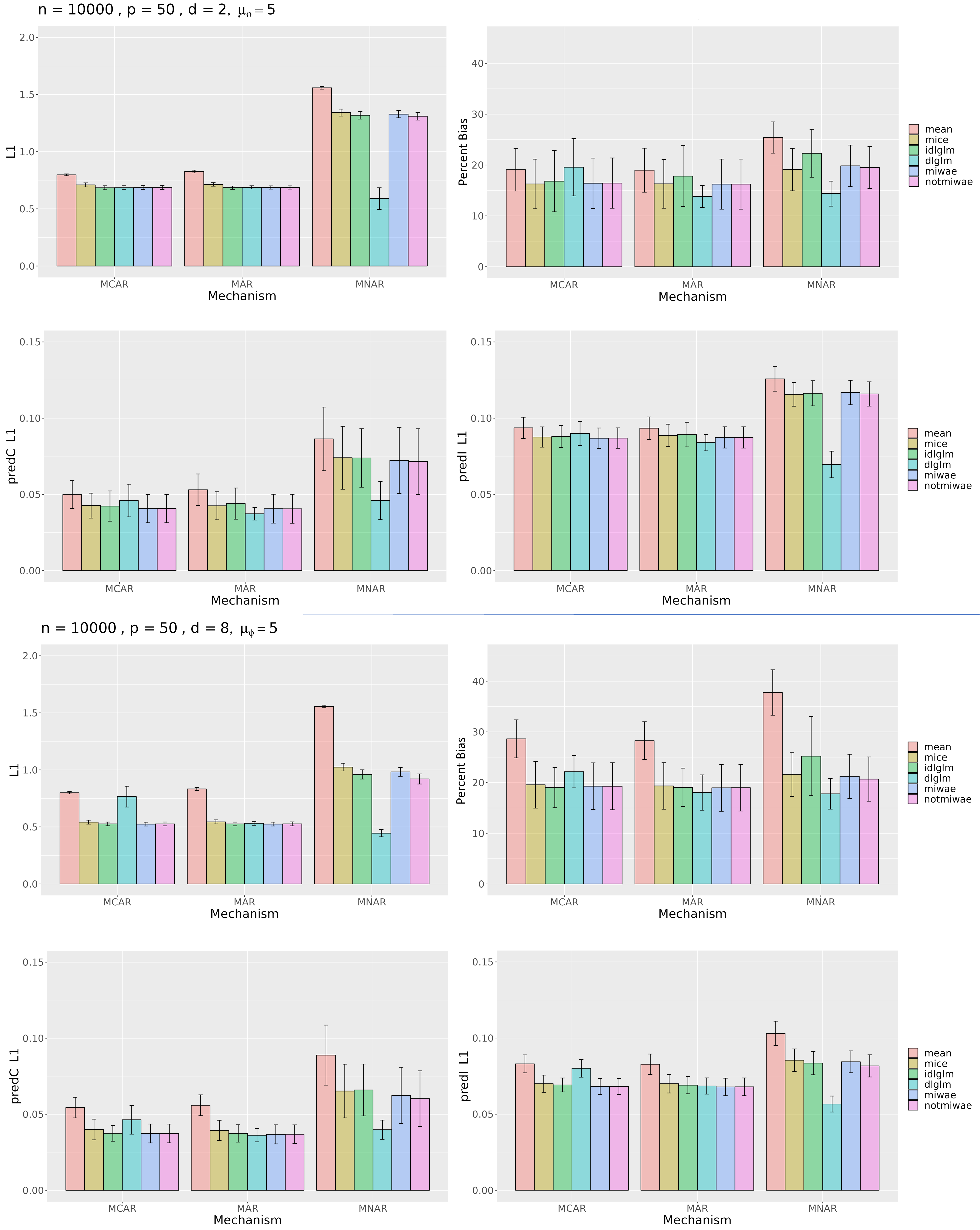}
\end{center}
\caption{\scriptsize Simulation results with $n=10,000$ and $p=50$, varying $d=2$ (top 4) and $d=8$ (bottom 4). In each quadrant, we measure imputation accuracy by the average L1 distance between imputed vs true values in $\mathbf{X}$ (top-left), coefficient estimation accuracy by the average percent bias (PB) of the estimates $\hat{\boldsymbol{\beta}}$ compared to the truth (top-right), and prediction accuracy by the average L1 distance between the predicted and true probabilities of class 1 membership of $\mathbf{Y}$ using the true unmasked test set (predC, bottom-left) and the incomplete test set (predI, bottom-right). In predI, we first impute missing values of the test data for mean, \textit{miwae}, \textit{notmiwae}, and \textit{mice} imputation, and we input the incomplete test set as-is for \textit{dlglm} and \textit{idlglm}.}
\label{fig:dlglmsimsN10KP50}
\end{figure}

\begin{figure}
\begin{center}
\includegraphics[width=155mm]{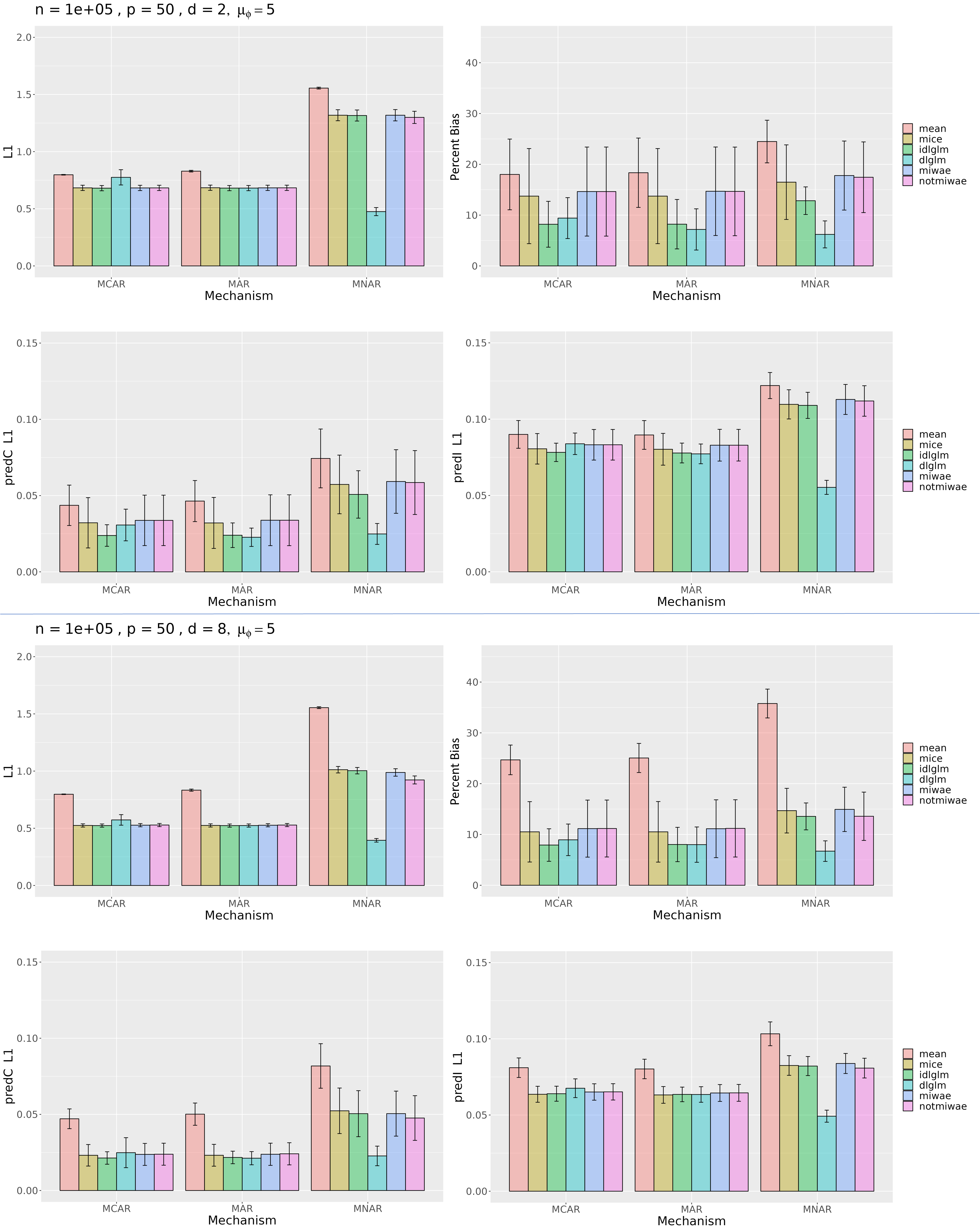}
\end{center}
\caption{\scriptsize Simulation results with $n=100,000$ and $p=50$, varying $d=2$ (top 4) and $d=8$ (bottom 4). In each quadrant, we measure imputation accuracy by the average L1 distance between imputed vs true values in $\mathbf{X}$ (top-left), coefficient estimation accuracy by the average percent bias (PB) of the estimates $\hat{\boldsymbol{\beta}}$ compared to the truth (top-right), and prediction accuracy by the average L1 distance between the predicted and true probabilities of class 1 membership of $\mathbf{Y}$ using the true unmasked test set (predC, bottom-left) and the incomplete test set (predI, bottom-right). In predI, we first impute missing values of the test data for mean, \textit{miwae}, \textit{notmiwae}, and \textit{mice} imputation, and we input the incomplete test set as is for \textit{dlglm} and \textit{idlglm}.}
\label{fig:dlglmsimsN100KP50}
\end{figure}

Figures \ref{fig:dlglmsimsN10KP50} and \ref{fig:dlglmsimsN100KP50} illustrate the simulation results pertaining to imputation accuracy, coefficient estimation, and prediction accuracy for the condition $p=50$. We see that across all combinations of $\{n,d\}$ and mechanisms of missingness, mean imputation consistently performs poorly in imputation, coefficient estimation, and prediction, while \textit{mice} and \textit{idlglm} perform comparably in these metrics. Also, we note that under MNAR missingness, \textit{dlglm} generally yields the lowest imputation and prediction error, as well as percent bias across all simulation cases. Under MAR missingness, \textit{dlglm} performs comparably to \textit{idlglm} and \textit{mice}. This shows the ability of \textit{dlglm} to learn an accurate model of the missingness, even under severe overparametrization of the missingness model (model need not be specified for ignorable missingness). However, due to the complexity of the model, we see that \textit{dlglm} does generally perform poorly compared to \textit{idlglm} and \textit{mice} under MCAR missingness, when $n=10,000$, although it still performs comparably to other methods when the sample size is very large ($n=100,000$). As one may expect, prediction performance using the incomplete data (predI) was poorer than prediction performance using the complete data (predC) for all methods.

We additionally show results pertaining to $p=25$ in Appendix B2 of the supplementary materials. We similarly found that \textit{dlglm} performed best under MNAR missingness, and comparably to \textit{idlglm}, \textit{mice}, \textit{miwae}, and \textit{notmiwae} under MCAR and MAR missingness.

\subsection{Real Data with Simulated Missingness} \label{sec:real3}

Next, we analyzed 3 completely-observed, large datasets from the UCI Machine Learning Repository \citep{Dua2017} that contained a specific response variable of interest, in order to preserve non-linearity and interactions between observed features. Unlike the simulated datasets, these UCI datasets don't follow a specific distribution that may be leveraged to inform a supervised learning method. The DRYBEAN dataset contains 16 features describing 13,611 images of dry beans taken with a high-resolution camera, and the response variable of interest was the type of dry bean each image represents, with 7 different possible types of beans. The LETTER dataset contains 16 attributes of 20,000 black-and-white pixel images, each displaying one English letter (A to Z). Finally, the SHUTTLE dataset contains 9 numerical attributes pertaining to 58,000 shuttle stat logs (observations), which are classified into 7 different categories. Due to a low sample size in 4 of the 7 categories, we pre-filtered the observations pertaining to these categories out of the dataset, and the resulting dataset contained 57,756 observations of 3 categories. In each of these datasets, the response variable was categorical with greater than two levels. Additional information regarding these datasets, and how to obtain the raw data files can be found in Appendix C of the supplementary materials.

We then simulated the missingness mask $\mathbf{R}^X$ with MCAR, MAR, and MNAR patterns of missingness in the  manner described in Section \ref{sec:simsetup3}. We split the samples in each dataset by a similar 8:1:1 ratio of training/validation/test samples.  In the test set samples, we then imputed the missing values and predicted the response variables with each method in a manner similar to Section \ref{sec:simsetup3}. For \textit{dlglm} and \textit{idlglm}, We account for potential nonlinear relationships between the covariates and response by allowing the number of hidden layers in $s_{\boldsymbol{\beta},\pi}(\cdot)$ to be nonzero in hyperparameter tuning. Then, we compared imputation and prediction accuracy on each dataset, under each mechanism of missingness. Since the underlying true probabilities of class membership were unavailable, we measured prediction accuracy by the Cohen's kappa metric on the complete (kappaC) and incomplete (kappaI) test set, with predicted class determined by the maximum predicted probability of membership. This metric measures how accurately a categorical variable was predicted, with a value of -1 indicating worst possible performance, and a value of 1 indicating perfect concordance with the truth.



\begin{sidewaysfigure}
\begin{center}
\includegraphics[width=205mm]{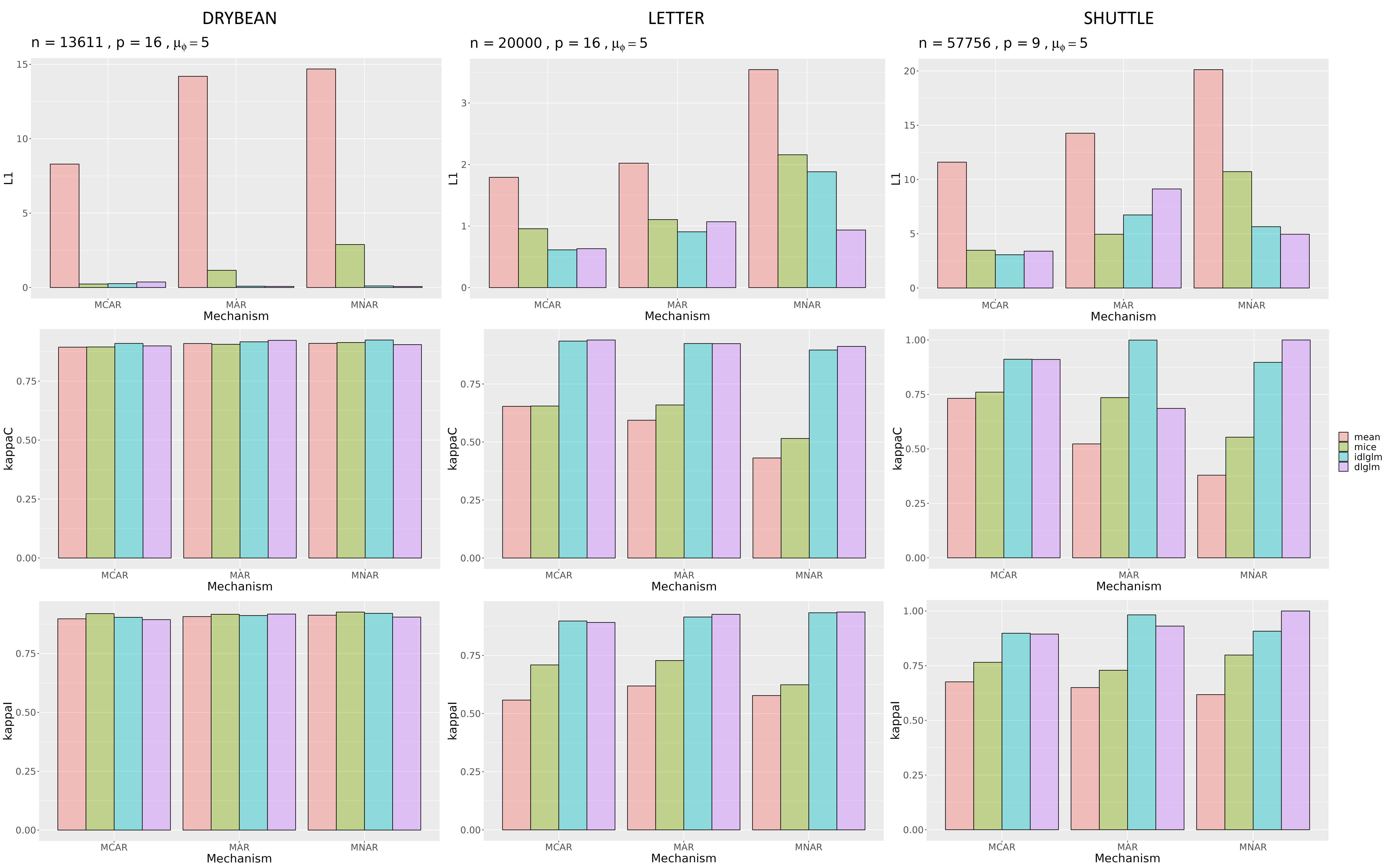}
\end{center}
\caption{\scriptsize Imputation (top row) and prediction results from predC (middle row) and predI (bottom row) from comparative methods run on 3 large datasets from the UCI Machine Learning Repository: DRYBEAN, LETTER, and SHUTTLE (columns, left to right). Imputation error was measured by the average L1 distance between true and imputed entries, with lower values indicating better performance, and prediction performance was measured by the Cohen's kappa metric for both predC (kappaC) and predI (kappaI), with higher values indicating better performance.}
\label{fig:dlglmUCI}
\end{sidewaysfigure}

Results from the imputation and prediction analyses on these datasets can be found in Figure \ref{fig:dlglmUCI}. We found that, as in the simulations, mean imputation performed most poorly in both imputation and downstream prediction, while \textit{dlglm} tended to perform best in the MNAR cases, and performed comparably to \textit{mice} and \textit{idlglm} under the MCAR and MAR cases. This further validates our claims under a more realistic setting, where the true data generation mechanism may be unknown. We also see that under both MCAR and MAR missingness, \textit{mice} performed worse than \textit{idlglm} in prediction on the LETTER and SHUTTLE datasets. The \textit{mice} model has been known to break down under nonlinear relationships between the features \citep{VanBuuren2018}, as may be the case in real-world datasets like the ones being examined. Using neural networks to model the data generation process allows \textit{idlglm} to better model potential nonlinear relationships between features, allowing for more accurate prediction.

Interestingly, all of the algorithms performed similarly in prediction on the DRYBEAN dataset. We found that this dataset contained extremely high levels of correlation between the variables (see Web Appendix C of the supplementary materials). When features containing missingness are highly correlated to other fully-observed features, such missingness may not truly reflect the MNAR scenario \citep{Hapfelmeier2012}. This is because there exist fully-observed features that are highly correlated with the missing features, and ignorably-missing data methods like \textit{idlglm} may gather information about the missing entries from these correlated, fully-observed features without having to explicitly model the mechanism of missingness. Still, \textit{idlglm} and \textit{dlglm} imputed missing entries much more accurately than \textit{mean} and \textit{mice} under MAR and MNAR. Interestingly, we also see that \textit{idlglm} performed similarly to \textit{dlglm} under MNAR in this dataset. 

We additionally performed similar analyses on 5 other smaller UCI datasets, and these results can be found in Appendix D of the supplementary materials. We found that under small sample size settings, performance via \textit{dlglm} was more variable under MNAR. We suggest use of \textit{dlglm} when the data contains at least 10,000 samples, as the model may be too complex to be accurately trained under smaller sample sizes.

\subsection{Bank Marketing Dataset} \label{sec:bankmarketing}

Finally, we performed prediction on the Bank Marketing dataset from the UCI Machine Learning Repository. This dataset contained 41,188 observations of 20 different attributes that were obtained based on direct phone calls from a Portuguese banking institution as part of a promotion campaign for a term deposit subscription \citep{Moro2014}. The response variable of interest was a fully-observed binary measure of whether the client subscribed a term deposit. Of the 20 attributes, we removed 1 attribute as directed from the manual due to perfect correlation with the response variable, and removed 3 other attributes that were deemed irrelevant to the prediction task: month of contact, day of contact, and communication type (cell phone or telephone).

Missingness was present in 8 of the 16 attributes: type of job, marital status, level of education, whether the client had a credit in default, whether the client had a housing loan, whether the client had a personal loan, number of days since the client was contacted in a previous campaign, and outcome of the previous campaign. The remaining 8 attributes were fully-observed: age of client, number of contacts during this campaign, number of contacts  before this campaign, employment variation rate, consumer price index, consumer confidence index, euribor 3 month rate, and employee number. The global rate of missingness was about 13.3\%. The response variable of interest was collected by additional follow-up calls to confirm whether the client subscribed to the product. Additional information regarding the bank marketing dataset, and how to obtain the raw data files can be found in Appendix E of the supplementary materials.

This type of dataset reflects the most realistic situation in practice, where missingness exists in a dataset and one has no prior knowledge of either the relationships between the features and the response, or the underlying mechanism of the missingness. We divided the dataset into the 8:1:1 training, validation, and test set ratio, and performed prediction as before. Because neither the data nor the missingness was simulated, we compared just the predI prediction performance across the methods.

In order to more deeply dive into this real data example, we assessed the prediction performance for dlglm and idlglm in the context of prediction (excluding $\mathbf{Y}$ from neural networks $g_{\theta_2}(\cdot)$ and $h_\phi(\cdot)$) and imputation (including $\mathbf{Y}$, denoted by dlglm$_y$ and idlglm$_y$).

\begin{table}[ht]
\centering
\begin{tabular}{lllll}
  \hline
 & AUC & PPV & kappa & F1 \\ 
  \hline
dlglm & 0.778 & 0.475 & 0.397 & 0.470 \\ 
\rowcolor[gray]{0.90} dlglm$_y$ & \textcolor{red}{0.880} & \textcolor{red}{0.481} & \textcolor{red}{0.445} & \textcolor{red}{0.516} \\ 
   idlglm & 0.791 & 0.475 & 0.407 & 0.460 \\ 
   \rowcolor[gray]{0.90}idlglm$_y$ & 0.779 & 0.446 & 0.411 & 0.488 \\ 
  mean & 0.769 & 0.448 & 0.385 & 0.46 \\ 
   \rowcolor[gray]{0.90}mice & 0.771 & 0.455 & 0.396 & 0.471 \\ 
   \hline
\end{tabular}
\caption{\scriptsize Results from prediction analyses on the Bank Marketing dataset from the UCI Machine Learning Repository. We measured concordance between the true and predicted binary response by 4 metrics: Area Under ROC Curve (AUC), Positive Predictivity (PPV), Cohen's kappa (kappa), and F1 score. For dlglm$_y$ and idlglm$_y$, $\mathbf{Y}$ was included in the input for neural networks $g_{\theta_2}(\cdot)$ and $h_\phi(\cdot)$, in order to allow for more accurate imputed values.} 
\label{tab:bankmarketing}
\end{table}

Table \ref{tab:bankmarketing} shows the results from these prediction analyses. We measured prediction performance of the binary response variable by 4 metrics: Area Under the ROC Curve (AUC), Positive Predictivity (PPV), Cohen's kappa (kappa), and the F1 metric. The formulas for PPV and F1 metrics are given in Appendix F of the supplementary materials. For each metric, a larger value represents greater concordance between the true and predicted response. We see that although \textit{dlglm}$_y$ yielded a significantly greater performance in prediction via all metrics, \textit{dlglm} does not significantly outperform \textit{idlglm}. The similar performance between \textit{dlglm} and \textit{idlglm} may indicate that the real mechanism of missingness in this data may not be MNAR, although this claim is not testable in practice.

Additionally, the trained \textit{dlglm} model chose 0 hidden layers in the $s_{\boldsymbol{\beta}, \boldsymbol{\pi}}(\mathbf{X})$ neural network in the optimal model, such that $s_{\boldsymbol{\beta}, \boldsymbol{\pi}}(\mathbf{X}) \rightarrow s_{\boldsymbol{\beta}}(\mathbf{X})$. Therefore, the weights of that neural network $\boldsymbol{\beta}$ coincide exactly with the coefficient estimates of a classic generalized linear model, i.e. $\text{logit}[Pr(Y=1)] = \mathbf{X}\boldsymbol{\beta}$. The features in the dataset with the largest effects on the probability of a client subscribing to a term deposit were employment variation rate (0.538), age of client (0.508), and whether the client had a personal loan (-0.402). Specifically, a client was more likely to subscribe if the company experienced higher levels of variation in employment and if the client were older, while a client was less likely to subscribe if they had a personal loan.

\section{Discussion} \label{sec:disc3}
In this paper, we introduced a novel deep learning method called Deeply-learned Generalized Linear Model with Missing Data (\textit{dlglm}), which is able to perform coefficient estimation and prediction in the presence of missing not at random (MNAR) data. \textit{dlglm} utilizes a deep learning neural network architecture to model the generation of the data matrix $\mathbf{X}$, as well as the relationships between the response variable $\mathbf{Y}$ and $\mathbf{X}$ and between the missingness mask $\mathbf{R}$ and $\mathbf{X}$. In this way, we are able to (1) generalize the traditional GLM to account for complex nonlinear interactions between the features, and (2) account for ignorable and non-ignorable forms of missingness in the data. We also showed through simulations and real data analyses that \textit{dlglm} performs better in coefficient estimation and prediction in the presence of MNAR missingness than other impute-then-regress methods, like mean and mice imputation. Furthermore, we found that \textit{dlglm} was generally robust to the mechanism of missingness, performing comparably well to \textit{mice} and \textit{idlglm} under MCAR and MAR settings. Still, it is recommended to utilize \textit{idlglm} when assuming the missingness is ignorable, given that the missingness model that is learned in \textit{dlglm} is not necessary in this setting.

Supervised learning algorithms such as \textit{dlglm} and \textit{idlglm} can be particularly useful in analyzing real-life data in the presence of missingness. In reality, the mechanism underlying missing values cannot be explicitly known or tested, but \textit{dlglm} may allow flexibility to evaluate multiple assumptions regarding the missingness mechanism. Furthermore, whereas impute-then-regress methods may typically require fully-observed observations at test time for prediction, \textit{dlglm} and \textit{idlglm} can predict the response of interest using partially-observed observations. This provides a convenient workflow, where a user need not separately re-impute the prediction set at test time.

In this paper, we focused specifically on the case of univariate response $\mathbf{Y}$. \textit{dlglm} and \textit{idlglm} can be generalized to the multivariate \textbf{Y} case by (1) including $\mathbf{Y}$ in the existing IWAE structure and (2) expanding the neural network $s_{\boldsymbol{\beta},\pi}(\mathbf{x}_i)$ to account for all $q$ responses in $\mathbf{Y}$, and utilizing samples of $\mathbf{Z}$ as additional input into this network such that $s_{\boldsymbol{\beta},\pi}(\mathbf{x}_i) \rightarrow s_{\boldsymbol{\beta},\pi}(\mathbf{x}_i, \mathbf{z}_{i})$. By doing this, we account for multivariate $\mathbf{Y}$, outputting additional parameters pertaining to the newly-specified distribution of $p_{\boldsymbol{\beta},\pi}(\mathbf{y}_i|\mathbf{x}_i,\mathbf{z}_i)$ and modelling correlation of \textbf{Y} by the learned latent structure. We leave this as an extension of our method.



\bigskip
\begin{center}
{\large\bf SUPPLEMENTARY MATERIAL}
\end{center}

\begin{description}
\item[Supplementary Materials:] Additional details of the \textit{dlglm} algorithm and the datasets used in this paper. (pdf)
\item[R-package for dlglm:] R-package \textit{dlglm} containing code to perform the diagnostic methods described in the article. The package can be downloaded from \url{https://github.com/DavidKLim/dlglm} (website)
\item[R Paper repo for reproducibility:] Github repository to replicate all analyses from this paper can be found here: \url{https://github.com/DavidKLim/dlglm_Paper} (website)
\end{description}



\bibliographystyle{agsm}
\bibliography{dlglm_JCGS}

@inproceedings{Li2020,
  title={ACFlow: Flow models for arbitrary conditional likelihoods},
  author={Li, Yang and Akbar, Shoaib and Oliva, Junier},
  booktitle={International Conference on Machine Learning},
  pages={5831--5841},
  year={2020},
  organization={PMLR}
}

@article{Strauss2021,
  title={Arbitrary conditional distributions with energy},
  author={Strauss, Ryan and Oliva, Junier B},
  journal={Advances in Neural Information Processing Systems},
  volume={34},
  pages={752--763},
  year={2021}
}

@article{Strauss2022,
  title={Posterior Matching for Arbitrary Conditioning},
  author={Strauss, Ryan and Oliva, Junier B},
  journal={Advances in Neural Information Processing Systems},
  volume={35},
  pages={18088--18099},
  year={2022}
}

@Article{Lim2021,
  author        = {{Lim}, David K. and {Rashid}, Naim U. and {Oliva}, Junier B. and {Ibrahim}, Joseph G.},
  title         = {{Handling Non-ignorably Missing Features in Electronic Health Records Data Using Importance-Weighted Autoencoders}},
  eid           = {arXiv:2101.07357},
  eprint        = {2101.07357},
  pages         = {arXiv:2101.07357},
  adsurl        = {https://ui.adsabs.harvard.edu/abs/2021arXiv210107357L},
  archiveprefix = {arXiv},
  journal       = {arXiv e-prints},
  month         = jan,
  primaryclass  = {cs.LG},
  year          = {2021},
}

@Article{Hapfelmeier2012,
  author    = {Alexander Hapfelmeier and Torsten Hothorn and Kurt Ulm and Carolin Strobl},
  title     = {A new variable importance measure for random forests with missing data},
  doi       = {10.1007/s11222-012-9349-1},
  number    = {1},
  pages     = {21--34},
  volume    = {24},
  journal   = {Statistics and Computing},
  month     = {aug},
  publisher = {Springer Science and Business Media {LLC}},
  year      = {2012},
}

@book{VanBuuren2018,
  title={Flexible imputation of missing data},
  author={Van Buuren, Stef},
  year={2018},
  publisher={CRC press}
}

@Article{VanBuuren2011,
  author  = {Van Buuren, Stef and Groothuis-Oudshoorn, Karin},
  title   = {mice: Multivariate imputation by chained equations in R},
  number  = {1},
  pages   = {1--67},
  volume  = {45},
  journal = {Journal of statistical software},
  year    = {2011},
}

@Article{Stubbendick2003,
  author    = {Amy L. Stubbendick and Joseph G. Ibrahim},
  title     = {Maximum Likelihood Methods for Nonignorable Missing Responses and Covariates in Random Effects Models},
  journal   = {Biometrics},
  year      = {2003},
  volume    = {59},
  number    = {4},
  month     = {dec},
  pages     = {1140--1150},
  doi       = {10.1111/j.0006-341x.2003.00131.x},
  publisher = {Wiley},
}

@article{Moro2014,
  title={A data-driven approach to predict the success of bank telemarketing},
  author={Moro, S{\'e}rgio and Cortez, Paulo and Rita, Paulo},
  journal={Decision Support Systems},
  volume={62},
  pages={22--31},
  year={2014},
  publisher={Elsevier}
}

@Article{Hoogland2020,
  author    = {Jeroen Hoogland and Marit Barreveld and Thomas P. A. Debray and Johannes B. Reitsma and Tom E. Verstraelen and Marcel G. W. Dijkgraaf and Aeilko H. Zwinderman},
  title     = {Handling missing predictor values when validating and applying a prediction model to new patients},
  doi       = {10.1002/sim.8682},
  number    = {25},
  pages     = {3591--3607},
  volume    = {39},
  journal   = {Statistics in Medicine},
  month     = {jul},
  publisher = {Wiley},
  year      = {2020},
}

@Article{Wells2013,
  author    = {Brian J. Wells and Amy S. Nowacki and Kevin Chagin and Michael W. Kattan},
  title     = {Strategies for Handling Missing Data in Electronic Health Record Derived Data},
  doi       = {10.13063/2327-9214.1035},
  number    = {3},
  pages     = {7},
  volume    = {1},
  journal   = {{eGEMs} (Generating Evidence {\&}: Methods to improve patient outcomes)},
  month     = {dec},
  publisher = {Ubiquity Press, Ltd.},
  year      = {2013},
}

@inproceedings{Guo1990,
  title={Analysis of gradient descent learning algorithms for multilayer feedforward neural networks},
  author={Guo, Heng and Gelfand, Saul B},
  booktitle={29th IEEE Conference on Decision and Control},
  pages={1751--1756},
  year={1990},
  organization={IEEE}
}

@article{Svozil1997,
  title={Introduction to multi-layer feed-forward neural networks},
  author={Svozil, Daniel and Kvasnicka, Vladimir and Pospichal, Jiri},
  journal={Chemometrics and intelligent laboratory systems},
  volume={39},
  number={1},
  pages={43--62},
  year={1997},
  publisher={Elsevier}
}

@Book{Rubin2004,
  author    = {Rubin, Donald B},
  title     = {Multiple imputation for nonresponse in surveys},
  publisher = {John Wiley \& Sons},
  volume    = {81},
  year      = {2004},
}

@article{Boyd2011,
  title={Distributed optimization and statistical learning via the alternating direction method of multipliers},
  author={Boyd, Stephen and Parikh, Neal and Chu, Eric and Peleato, Borja and Eckstein, Jonathan and others},
  journal={Foundations and Trends{\textregistered} in Machine learning},
  volume={3},
  number={1},
  pages={1--122},
  year={2011},
  publisher={Now Publishers, Inc.}
}

@inproceedings{Ghorbani2018,
  title={Embedding for informative missingness: Deep learning with incomplete data},
  author={Ghorbani, Amirata and Zou, James Y},
  booktitle={2018 56th Annual Allerton Conference on Communication, Control, and Computing (Allerton)},
  pages={437--445},
  year={2018},
  organization={IEEE}
}

@article{Lydia2019,
  title={Adagrad—an optimizer for stochastic gradient descent},
  author={Lydia, Agnes and Francis, Sagayaraj},
  journal={Int. J. Inf. Comput. Sci},
  volume={6},
  number={5},
  pages={566--568},
  year={2019}
}

@WWW{Dormehl2019,
  author = {Luke Dormehl},
  title  = {What is an artificial neural network? Here’s everything you need to know},
  year   = {2019},
  date   = {2019-01-05},
  url    = {https://www.digitaltrends.com/cool-tech/what-is-an-artificial-neural-network/},
}

@article{Jang2016,
  title={Categorical reparameterization with gumbel-softmax},
  author={Jang, Eric and Gu, Shixiang and Poole, Ben},
  journal={arXiv preprint arXiv:1611.01144},
  year={2016}
}

@article{Lipsitz1996,
    author = {Lipsitz, Stuart R. and Ibrahim, Joseph G.},
    title = "{A conditional model for incomplete covariates in parametric regression models}",
    journal = {Biometrika},
    volume = {83},
    number = {4},
    pages = {916-922},
    year = {1996},
    month = {12},
    issn = {0006-3444},
    doi = {10.1093/biomet/83.4.916},
    eprint = {https://academic.oup.com/biomet/article-pdf/83/4/916/703810/83-4-916.pdf},
}

@incollection{Bottou2012,
  title={Stochastic gradient descent tricks},
  author={Bottou, L{\'e}on},
  booktitle={Neural networks: Tricks of the trade},
  pages={421--436},
  year={2012},
  publisher={Springer}
}

@article{Qi2003,
title = {Nonlinear prediction of exchange rates with monetary fundamentals},
journal = {Journal of Empirical Finance},
volume = {10},
number = {5},
pages = {623-640},
year = {2003},
issn = {0927-5398},
doi = {https://doi.org/10.1016/S0927-5398(03)00008-2},
author = {Min Qi and Yangru Wu}
}

@article{Nelder1972,
 ISSN = {00359238},
 author = {J. A. Nelder and R. W. M. Wedderburn},
 journal = {Journal of the Royal Statistical Society. Series A (General)},
 number = {3},
 pages = {370--384},
 publisher = {[Royal Statistical Society, Wiley]},
 title = {Generalized Linear Models},
 urldate = {2022-06-22},
 volume = {135},
 year = {1972}
}

@Article{Friedman2010,
  author    = {Jerome Friedman and Trevor Hastie and Robert Tibshirani},
  title     = {Regularization Paths for Generalized Linear Models via Coordinate Descent},
  journal   = {Journal of Statistical Software},
  year      = {2010},
  volume    = {33},
  number    = {1},
  doi       = {10.18637/jss.v033.i01},
  publisher = {Foundation for Open Access Statistic},
}

@article{Holland1977,
author = { Paul W.   Holland  and  Roy E.   Welsch },
title = {Robust regression using iteratively reweighted least-squares},
journal = {Communications in Statistics - Theory and Methods},
volume = {6},
number = {9},
pages = {813-827},
year  = {1977},
publisher = {Taylor & Francis},
doi = {10.1080/03610927708827533}

}

@book{Mccullagh2019,
  title={Generalized linear models},
  author={McCullagh, Peter and Nelder, John A},
  year={2019},
  publisher={Routledge}
}

@inproceedings{Ipsen2021,
  title={How to deal with missing data in supervised deep learning?},
  author={Ipsen, Niels Bruun and Mattei, Pierre-Alexandre and Frellsen, Jes},
  booktitle={International Conference on Learning Representations},
  year={2021}
}

@article{Ma2021,
  title={Identifiable generative models for missing not at random data imputation},
  author={Ma, Chao and Zhang, Cheng},
  journal={Advances in Neural Information Processing Systems},
  volume={34},
  pages={27645--27658},
  year={2021}
}

@article{Beesley2019,
  title={Sequential imputation for models with latent variables assuming latent ignorability},
  author={Beesley, Lauren J and Taylor, Jeremy MG and Little, Roderick JA},
  journal={Australian \& New Zealand Journal of Statistics},
  volume={61},
  number={2},
  pages={213--233},
  year={2019},
  publisher={Wiley Online Library}
}

@Misc{Dua2017,
  author      = {Dua, Dheeru and Graff, Casey},
  title       = {{UCI} Machine Learning Repository},
  year        = {2017},
  url         = {http://archive.ics.uci.edu/ml},
  institution = {University of California, Irvine, School of Information and Computer Sciences},
}

@Article{Diggle1994,
  author    = {P. Diggle and M. G. Kenward},
  title     = {Informative Drop-Out in Longitudinal Data Analysis},
  journal   = {Applied Statistics},
  year      = {1994},
  volume    = {43},
  number    = {1},
  pages     = {49},
  doi       = {10.2307/2986113},
  publisher = {{JSTOR}},
}

@incollection{Prechelt1998,
  title={Early stopping-but when?},
  author={Prechelt, Lutz},
  booktitle={Neural Networks: Tricks of the trade},
  pages={55--69},
  year={1998},
  publisher={Springer}
}

@InProceedings{Gershman2014,
  author    = {Samuel J. Gershman and Noah D. Goodman},
  title     = {Amortized Inference in Probabilistic Reasoning},
  booktitle = {CogSci},
  year      = {2014},
}

@Article{Cremer2017,
  author        = {{Cremer}, Chris and {Morris}, Quaid and {Duvenaud}, David},
  title         = {{Reinterpreting Importance-Weighted Autoencoders}},
  journal       = {arXiv e-prints},
  year          = {2017},
  eid           = {arXiv:1704.02916},
  month         = {Apr},
  pages         = {arXiv:1704.02916},
  eprint        = {1704.02916},
  adsurl        = {https://ui.adsabs.harvard.edu/abs/2017arXiv170402916C},
  archiveprefix = {arXiv},
  primaryclass  = {stat.ML},
}

@Article{Burda2015,
  author        = {{Burda}, Yuri and {Grosse}, Roger and {Salakhutdinov}, Ruslan},
  title         = {{Importance Weighted Autoencoders}},
  journal       = {arXiv e-prints},
  year          = {2015},
  eid           = {arXiv:1509.00519},
  month         = {Sep},
  pages         = {arXiv:1509.00519},
  eprint        = {1509.00519},
  adsurl        = {https://ui.adsabs.harvard.edu/abs/2015arXiv150900519B},
  archiveprefix = {arXiv},
  primaryclass  = {cs.LG},
}

@Article{Tran2019,
  author    = {M.-N. Tran and N. Nguyen and D. Nott and R. Kohn},
  title     = {Bayesian Deep Net {GLM} and {GLMM}},
  doi       = {10.1080/10618600.2019.1637747},
  number    = {1},
  pages     = {97--113},
  volume    = {29},
  journal   = {Journal of Computational and Graphical Statistics},
  month     = {aug},
  publisher = {Informa {UK} Limited},
  year      = {2019},
}

@Article{Ibrahim2009,
  author    = {Joseph G. Ibrahim and Geert Molenberghs},
  title     = {Missing data methods in longitudinal studies: a review},
  journal   = {{TEST}},
  year      = {2009},
  volume    = {18},
  number    = {1},
  month     = {feb},
  pages     = {1--43},
  doi       = {10.1007/s11749-009-0138-x},
  publisher = {Springer Science and Business Media {LLC}},
}

@Article{Ibrahim2005,
  author    = {Joseph G Ibrahim and Ming-Hui Chen and Stuart R Lipsitz and Amy H Herring},
  title     = {Missing-Data Methods for Generalized Linear Models},
  journal   = {Journal of the American Statistical Association},
  year      = {2005},
  volume    = {100},
  number    = {469},
  month     = {mar},
  pages     = {332--346},
  doi       = {10.1198/016214504000001844},
  publisher = {Informa {UK} Limited},
}

@Book{Little2002,
  author    = {Roderick J. A. Little and Donald B. Rubin},
  title     = {Statistical Analysis with Missing Data},
  year      = {2002},
  publisher = {John Wiley {\&} Sons, Inc.},
  doi       = {10.1002/9781119013563},
  month     = {aug},
}

@INPROCEEDINGS{Saxe2014,
    author = {Andrew M. Saxe and James L. Mcclelland and Surya Ganguli},
    title = {Exact solutions to the nonlinear dynamics of learning in deep linear neural network},
    booktitle = {In International Conference on Learning Representations},
    year = {2014}
}

@article{Murphy2016,
  title={An overview of convolutional neural network architectures for deep learning},
  author={Murphy, John},
  journal={Microway Inc},
  pages={1--22},
  year={2016}
}

@InProceedings{Mattei2019,
  author    = {Mattei, Pierre-Alexandre and Frellsen, Jes},
  title     = {{MIWAE}: Deep Generative Modelling and Imputation of Incomplete Data Sets},
  booktitle = {Proceedings of the 36th International Conference on Machine Learning},
  year      = {2019},
  editor    = {Chaudhuri, Kamalika and Salakhutdinov, Ruslan},
  volume    = {97},
  series    = {Proceedings of Machine Learning Research},
  publisher = {PMLR},
  month     = {09--15 Jun},
  pages     = {4413--4423},
  address   = {Long Beach, California, USA},
}

@Article{Lopez2018,
  author    = {Romain Lopez and Jeffrey Regier and Michael B. Cole and Michael I. Jordan and Nir Yosef},
  title     = {Deep generative modeling for single-cell transcriptomics},
  journal   = {Nature Methods},
  year      = {2018},
  volume    = {15},
  number    = {12},
  month     = {nov},
  pages     = {1053--1058},
  doi       = {10.1038/s41592-018-0229-2},
  publisher = {Springer Science and Business Media {LLC}},
}

@InCollection{Razzak2017,
  author    = {Muhammad Imran Razzak and Saeeda Naz and Ahmad Zaib},
  title     = {Deep Learning for Medical Image Processing: Overview, Challenges and the Future},
  booktitle = {Lecture Notes in Computational Vision and Biomechanics},
  year      = {2017},
  publisher = {Springer International Publishing},
  pages     = {323--350},
  doi       = {10.1007/978-3-319-65981-7_12},
  month     = {nov},
}

@Article{Rubin1976,
  author    = {Donald B. Rubin},
  title     = {Inference and missing data},
  journal   = {Biometrika},
  year      = {1976},
  volume    = {63},
  number    = {3},
  pages     = {581--592},
  doi       = {10.1093/biomet/63.3.581},
  publisher = {Oxford University Press ({OUP})},
}

@Article{Chen2019,
  author    = {David Chen and Sijia Liu and Paul Kingsbury and Sunghwan Sohn and Curtis B. Storlie and Elizabeth B. Habermann and James M. Naessens and David W. Larson and Hongfang Liu},
  title     = {Deep learning and alternative learning strategies for retrospective real-world clinical data},
  journal   = {npj Digital Medicine},
  year      = {2019},
  volume    = {2},
  number    = {1},
  month     = {may},
  doi       = {10.1038/s41746-019-0122-0},
  publisher = {Springer Science and Business Media {LLC}},
}

@Article{Kingma2014,
  author        = {{Kingma}, Diederik P. and {Ba}, Jimmy},
  title         = {{Adam: A Method for Stochastic Optimization}},
  eid           = {arXiv:1412.6980},
  eprint        = {1412.6980},
  pages         = {arXiv:1412.6980},
  adsurl        = {https://ui.adsabs.harvard.edu/abs/2014arXiv1412.6980K},
  archiveprefix = {arXiv},
  journal       = {arXiv e-prints},
  month         = dec,
  primaryclass  = {cs.LG},
  year          = {2014},
}

@Article{Kingma2013,
  author        = {{Kingma}, Diederik P and {Welling}, Max},
  title         = {{Auto-Encoding Variational Bayes}},
  journal       = {arXiv e-prints},
  year          = {2013},
  eid           = {arXiv:1312.6114},
  month         = {Dec},
  pages         = {arXiv:1312.6114},
  eprint        = {1312.6114},
  adsurl        = {https://ui.adsabs.harvard.edu/abs/2013arXiv1312.6114K},
  archiveprefix = {arXiv},
  primaryclass  = {stat.ML},
}

@Article{Kingma2019,
  author        = {{Kingma}, Diederik P. and {Welling}, Max},
  title         = {{An Introduction to Variational Autoencoders}},
  journal       = {arXiv e-prints},
  year          = {2019},
  eid           = {arXiv:1906.02691},
  month         = {Jun},
  pages         = {arXiv:1906.02691},
  eprint        = {1906.02691},
  adsurl        = {https://ui.adsabs.harvard.edu/abs/2019arXiv190602691K},
  archiveprefix = {arXiv},
  primaryclass  = {cs.LG},
}

\end{document}


\maketitle

\setstretch{2}


\section{Appendix A: Additional Details}

\subsection{A1: Computational Details}\label{sec:comp}
In this section, we go through details of some of the computational methods involved in \textit{dlglm} and \textit{idlglm}.

\subsubsection{Stochastic Gradient Descent}
In most deep learning architectures, stochastic gradient descent (SGD) is the favored algorithm of optimization, as it is very scalable to higher dimensions. A typical SGD algorithm proceeds as follows: let $Q$ be an objective function to be maximized and let $\Omega$ denote the collection of all parameters one wishes to maximize over.  Also let $Q^{(t)}$ denote the value of the objective function and $\hat{\Omega}^{(t)}$ denoting the estimate of $\Omega$ at update step $t$. Then, one can optimize $\Omega$ with respect to $Q$ by some update rule $\hat{\Omega}^{(t+1)} = \hat{\Omega}^{(t)} + \delta \nabla_\Omega  Q^{(t)}, $ where $\delta$ is the step size which controls how large of a change is applied to the update, and the gradient $\nabla$ is taken with respect to each parameter in $\Omega$. In this way, $\Omega$ is updated at each step $t$ such that $Q^{(t)}$ is increased, and the magnitude of the change in $\hat{\Omega}^{(t)}$ is mediated by $\delta$.

There are many variants of SGD, including ADAM \citep{Kingma2014}, ADMM \citep{Boyd2011}, Adagrad \citep{Lydia2019}, and more, as well as a natural gradient variational approximation with factor covariance method as discussed by \citet{Tran2019}. In \textit{dlglm} and \textit{idlglm}, we utilize ADAM as the default optimizer.

\subsubsection{Reparameterization Trick}
In a VAE, one uses the so-called reparameterization trick in order to draw samples from any continuous distribution $q_\theta(\mathbf{Z}|\mathbf{X})$ while allowing for a Monte Carlo estimate of the expectation in the lower bound to be differentiable with respect to $\theta$ \citep{Kingma2013}. This is accomplished by expressing the random variable $\mathbf{Z}=g_\theta(\epsilon,\mathbf{X})$, where $\boldsymbol{\epsilon}$ is an auxiliary variable with marginal distribution $p(\boldsymbol{\epsilon})$ and $g_\theta(\cdot)$ is some function with parameters $\theta$. In a traditional VAE setting where $q_\theta(\mathbf{Z}|\mathbf{X})$ is a multivariate normal distribution with diagonal covariance structure, $\boldsymbol{\epsilon} \sim N(0,1)$, and $g_\theta(\epsilon,\mathbf{X}) = \boldsymbol{\mu}_Z + \boldsymbol{\sigma}_Z \cdot \boldsymbol{\epsilon}$, where $\boldsymbol{\mu}_Z$ is the vector of means of $q_\theta(\mathbf{Z}|\mathbf{X})$, and $\boldsymbol{\sigma}_Z$ is the corresponding entries of the standard deviations from the diagonal covariance of $q_\theta(\mathbf{Z}|\mathbf{X})$.

In \textit{dlglm} and \textit{idlglm}, we multiply use the reparameterization trick for each of the $K$ latent variable pairs $\{\mathbf{Z}_k,\mathbf{X}_k^m\}$. The standard reparameterization trick can be used when all of the covariates are continuous, such that a multivariate normal distributional assumption with diagonal covariance structure may hold for both $q_{\theta_1}(\mathbf{z}_i|\mathbf{x}_i^o)$ and $q_{\theta_2}(\mathbf{x}_i^m|\mathbf{z}_i,\mathbf{x}_i^o,\mathbf{r}_i^X)$.

If some partially-observed covariates are categorical features, one can approximate a discrete categorical distribution with the Gumbel-Softmax distribution \citep{Jang2016}. In particular, without loss of generality, let $\mathbf{X}^{*}$ be a vector of length $n$ matrix consisting of one categorical feature in $\mathbf{X}$ with $C$ distinct classes, with entries $x_1^{*},\ldots,x_n^{*}$. Denote $\{x_{i1}^{*},\ldots,x_{iC}^{*}\}$ as the softmax probabilities of the categorical value of $x_i^{*}$ for each class. Then, the density of the Gumbel-Softmax distribution is written as $$p_{\rho,\tau}(x_{i1}^{*},\ldots,x_{iC}^{*})=\Gamma(C)\tau^{C-1}(\Sigma_{c=1}^{C}\pi_c/(x_{ic}^{*})^\tau)^{-C}\prod_{c=1}^{C}(\pi_c/(x_{ic}^{*})^{\tau+1}),$$ where $\tau$ is the softmax temperature and $\rho$ is the $C$ class probabilities of $x_{i}^*$. Since this is a continuous distribution, samples can be drawn using the standard reparameterization trick. When $\tau>>0$, $\{x_{i1}^{*},\ldots,x_{iC}^{*}\}$ represent softmax samples that whose values are close to $1/C$. As $\tau \rightarrow 0$, samples drawn from $p_{\rho,\tau}(x_{i1}^{*},\ldots,x_{iC}^{*})$ become one-hot, and this Gumbel-Softmax distribution becomes exactly the discrete categorical distribution.

\subsubsection{Initialization}
Initialization of weights and biases in deep learning architectures are typically done by drawing values centered around 0, such that the magnitudes of these parameters start off small, and those pertaining to important connections are increase in magnitude as training progresses, while the weights of unimportant connections tend towards 0. \textit{dlglm} and \textit{idlglm} uses the semi-orthogonal matrix initialization \citep{Saxe2014} for $\pi$, $\boldsymbol{\beta}$ $\psi$, $\theta$, and $\phi$, where for each layer, a QR decomposition is performed on random values drawn from $N(0,1)$ in order to obtain an orthogonal matrix, whose entries are used as initialized values for the weights and bias pertaining to that layer. This procedure is repeated for each layer of each neural network.

An orthogonal matrix of initialized values for the weights and biases have the nice property of being norm-preserving, which has been shown to aid in deep neural networks to prevent exploding or vanishing signals during training \citep{Saxe2014}.

\subsubsection{Early Stop Criterion}

In \textit{dlglm} and \textit{idlglm}, we incorporate an early stop criterion \citep{Prechelt1998} in order to prevent overfitting on the training set, and to reduce computation time. Specifically, let $L^{(t)} \equiv \hat{\mathcal{L}}_{K,valid}^{dlglm, (t)}$ denote the estimated \textit{dlglm} bound on a held-out validation set at each update step $(t)$, and initialize $L_{opt} = L^{(0)}$ and $E^{(0)} = 0$. For $t\geq 1$, if $L^{(t)} - L_{opt} > 0$, we replace $L_{opt} = L^{(t)}$. Also, if $L^{(t)} - L_{opt} \leq \varepsilon L_{opt}$, we set $E^{(t)}=E^{(t-1)}+1$. In this way, training continues as long as the improvement in the estimated lower bound in the validation set is greater than $\varepsilon L_{opt}$, but if not, training is allowed to run for a set number of updates before early stopping. This leeway (called ``$patience$'') is allowed due to the properties of stochastic gradient descent, which may cause $L^{(t)}$ to fluctuate, especially for smaller $t$. If $E^{(t)} = patience$, we save the optimal model pertaining to $L_{opt}$, and terminate training. Here, we set $patience=50$ and $\varepsilon=0.0001$.

\subsection{A2: Lower Bound with Known Covariate Distribution}

In some cases, it may be more appropriate in \textit{dlglm} to model the covariate distribution $p(\mathbf{X})$ with a known closed-form distribution, as described in Section 2.4.1 of the main text, rather than the VAE/IWAE structure as described in Section 2.4.2 that is utilized by default in \textit{dlglm}. Importantly, we no longer assume that the covariates $\mathbf{X}$ are generated from some lower-dimensional latent variable $\mathbf{Z}$, as in a typical VAE/IWAE setting.

In this special case, a lower bound on the marginal log-likelihood in the presence of missingness can be computed as in Section 2.4.3. Under MNAR, this quantity, which we call $\mathcal{L}_K^{dlglmX}$, can be derived as
\begin{align*}
\log &p_{\alpha, \boldsymbol{\beta}, \pi, \psi, \phi}(\mathbf{X}^o,\mathbf{Y},\mathbf{R}^X) = \sum_{i=1}^{n} \log p_{\alpha, \boldsymbol{\beta}, \pi, \psi, \phi}(\mathbf{x}_i^o,y_i,\mathbf{r}_i^X) \\
&= \sum_{i=1}^n \log \left[\int p_{\alpha, \boldsymbol{\beta}, \pi, \psi, \phi}(\mathbf{x}_i^o,\mathbf{x}_i^m,y_i,\mathbf{r}_i^X) d\mathbf{x}_i^m \right]\\
&= \sum_{i=1}^n \log \E_{\mathbf{x}_{ik}^m \sim q_{\theta}(\mathbf{x}_i^m|\mathbf{x}_i^o,\mathbf{r}_i^X)} \left[ \frac{1}{K}\sum_{k=1}^K  \frac{p_{ \psi}(\mathbf{x}_i^o,\mathbf{x}_{ik}^m)p_{\alpha, \boldsymbol{\beta}, \pi}(y_i|\mathbf{x}_i^o,\mathbf{x}_{ik}^m)p_\phi(\mathbf{r}_i^X|\mathbf{x}_i^o,\mathbf{x}_{ik}^m)}{q_{\theta}(\mathbf{x}_{ik}^m|\mathbf{x}_i^o,\mathbf{r}_i^X)} \right]\\
&\geq \sum_{i=1}^{n} \E_{\mathbf{x}_{ik}^m \sim q_{\theta}(\mathbf{x}_i^m|\mathbf{x}_i^o,\mathbf{r}_i^X)} \log{ \left[ \frac{1}{K}\sum_{k=1}^{K}\frac{p_{ \psi}(\mathbf{x}_i^o,\mathbf{x}_{ik}^m)p_{\alpha, \boldsymbol{\beta}, \pi}(y_i|\mathbf{x}_i^o,\mathbf{x}_{ik}^m)p_\phi(\mathbf{r}_i^X|\mathbf{x}_i^o,\mathbf{x}_{ik}^m)}{q_{\theta}(\mathbf{x}_{ik}^m|\mathbf{x}_i^o,\mathbf{r}_i^X)} \right] } = \mathcal{L}_{K}^{dlglmX}.
\end{align*}
Similar to the default derivation of \textit{dlglm}, we assume $\mathbf{x}_{i1}^m,\ldots,\mathbf{x}_{iK}^m \stackrel{i.i.d}{\sim} q_{\theta}(\mathbf{x}_{i}^m|\mathbf{x}_i^o,\mathbf{r}_i^X)$, except now we have just the latent variables pertaining to the missing data, the weights and biases of the neural network that learns the variational posterior is now $\{\theta_1,\theta_2\} \rightarrow \theta$, and we need not specify a form for either the prior $p(\mathbf{z}_{ik})$ or the variational joint posterior which includes the the latent variables pertaining to the lower-dimensional representation $\mathbf{z}_{i1},\ldots,\mathbf{z}_{iK}$. Instead, we assume an explicit covariate distribution $p_\psi(\mathbf{x}_i^o,\mathbf{x}_{ik}^m)$, which is indexed by some parameters $\psi$. For example, if $p_\psi(\mathbf{x}_i^o,\mathbf{x}_{ik}^m)$ is assumed to be distributed multivariate normal with an independent covariance structure, then $\psi=\{\boldsymbol{\mu},\boldsymbol{\Sigma}\}$, where $\boldsymbol{\mu}$ is a mean vector of length $p$, and $\boldsymbol{\Sigma}$ is a $p \times p$ covariance matrix. One can then optimize values of $\psi$ in conjunction with the weights and biases of the neural network architecture via stochastic gradient descent during training.

Also, the architecture would not contain the encoder $f_{\psi}(\mathbf{z}_i)$ and decoder $g_{\theta_1}(\mathbf{x}_i^o)$ neural networks as in Section 2.4.3, but would consist of the neural networks $g_{\theta}(\mathbf{x}_i^o,\mathbf{r}_i^X)$, $s_{\boldsymbol{\beta}, \pi}(\mathbf{x}_i)$, and $h_{\phi}(\mathbf{x}_i)$, which would respectively output the parameters of $q_\theta(\mathbf{x}_i^m|\mathbf{x}_i^o,\mathbf{r}_i^X)$, $p(y_i|\mathbf{x}_i^o,\mathbf{x}_{ik}^m)$, and $p(\mathbf{r}_i^X|\mathbf{x}_i^o,\mathbf{x}_{ik}^m)$. Values of the missing entries will similarly be drawn from the $q_\theta(\mathbf{x}_i^m|\mathbf{x}_i^o,\mathbf{r}_i^X)$, whose parameters are output by the neural network $g_{\theta}(\mathbf{x}_i^o,\mathbf{r}_i^X)$, and the drawn values will be used as input for the downstream neural networks $s_{\boldsymbol{\beta}, \pi}(\mathbf{x}_i)$ and $h_{\phi}(\mathbf{x}_i)$.

Under ignorable missingness, the analogous lower bound can be similarly obtained as in Section 2.4.3 for \textit{idlglm} by assuming independence between the missing values $\mathbf{X}^m$ and the missingness mask $\mathbf{R}^X$: $$ \mathcal{L}_{K}^{idlglmX} = \sum_{i=1}^{n} \E_{\mathbf{x}_{ik}^m \sim q_{\theta}(\mathbf{x}_i^m|\mathbf{x}_i^o)} \log{ \left[ \frac{1}{K}\sum_{k=1}^{K}\frac{p_{ \psi}(\mathbf{x}_i^o,\mathbf{x}_{ik}^m)p_{\alpha, \boldsymbol{\beta}, \pi}(y_i|\mathbf{x}_i^o,\mathbf{x}_{ik}^m)p_\phi(\mathbf{r}_i^X|\mathbf{x}_i^o)}{q_{\theta}(\mathbf{x}_{ik}^m|\mathbf{x}_i^o)} \right] }$$

\subsection{A3: IWAE Architecture}
\begin{figure}[H]
\begin{center}
\includegraphics[width=130mm]{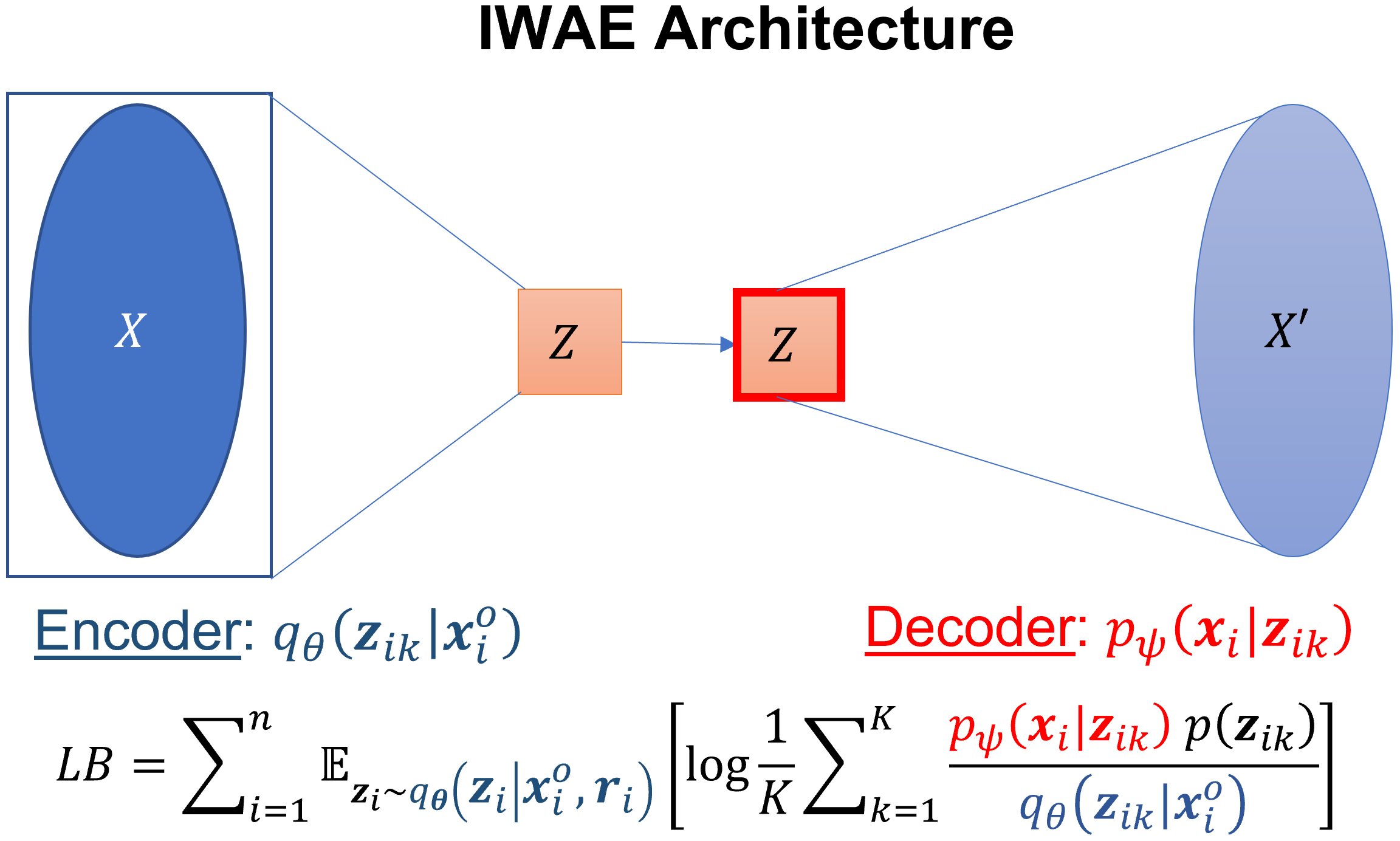} 
\end{center}
\caption{\small Architecture of an importance weighted autoencoder (IWAE) in the absence of missing data. Darkly colored nodes represent deterministic values, lightly colored nodes represent learned distributional parameters, and outlined (in red) nodes represent sampled values from learned distributions. Orange cells correspond to latent variables $\mathbf{Z}$. $\mathbf{Z}$ is sampled $K$ times from its posterior distribution Below is the lower bound ($LB$), which is optimized via stochastic gradient descent.}
\label{fig:IWAEarchitecture}
\end{figure}

\subsection{A4: dlglm Training Algorithm}

The training of the \textit{dlglm} architecture proceeds as follows:
\begin{enumerate}
\item The missing entries are pre-imputed to zero and appended with observed entries, and fed into $g_{\theta_1}(\mathbf{x}_i^o)$, to learn parameters of $q_{\theta_1}(\mathbf{z}_i|\mathbf{x}_i^o)$.
\item $K$ samples are drawn from $q_{\theta_1}(\mathbf{Z}|\mathbf{X}^o,\mathbf{R})$.
\item Samples from (2) are used as input for $f_\psi(\mathbf{z}_i)$, to learn the parameters of $p_{\psi}(\mathbf{x}_i|\mathbf{z}_i)$.
\item The samples from (2) are used again as input for $g_{\theta_2}(\mathbf{z}_i,\mathbf{r}_i^X,\mathbf{x}_i^o)$, concatenated with the observed data entries (with pre-imputed missing entries) and the missingness mask to learn parameters of $q_{\theta_2}(\mathbf{x}_i^m|\mathbf{z}_i,\mathbf{x}_i^o,\mathbf{r}_i^X)$.
\item We draw samples of $\mathbf{x}_i^m$ from $q_{\theta_2}(\mathbf{x}_i^m|\mathbf{z}_i,\mathbf{x}_i^o,\mathbf{r}_i^X)$, and use them as input, concatenated with the fixed observed entries $\mathbf{x}_i^o$, into $h_\phi(\mathbf{x}_i)$ (or the mask decoder network) to learn the parameters associated with the model of the missingness mask $p_\phi(\mathbf{r}_i^X|\mathbf{x}_i)$.
\item We use the samples of $\mathbf{x}_i^m$ from (5), concatenated with the observed variables $\mathbf{x}_i^o$ as input into $s_{\boldsymbol{\beta},\pi}(\mathbf{x}_i)$ to output parameters of $p_{\alpha,\boldsymbol{\beta},\pi}(y_i|\mathbf{x}_i)$. The dispersion parameter $\alpha$ is additionally learned via stochastic gradient descent, along with the all of the weights and biases of the entire architecture.
\end{enumerate}

Under simple distributional assumptions of $q_{\theta_2}(\mathbf{x}_i^m|\mathbf{z}_i,\mathbf{r}_i^X,\mathbf{x}_i^o)$, the sampling step in Step (5) is similar to the sampling of the latent variable $\mathbf{Z}$ in Step (2), and both can be accomplished using the reparametrization trick \citep{Kingma2013}. These steps outline the case where $\mathbf{R}^X$ is independent on $\mathbf{Z}$.

\subsection{A5: dlglm Case xy}
\begin{figure}[H]
\begin{center}
\includegraphics[width=165mm]{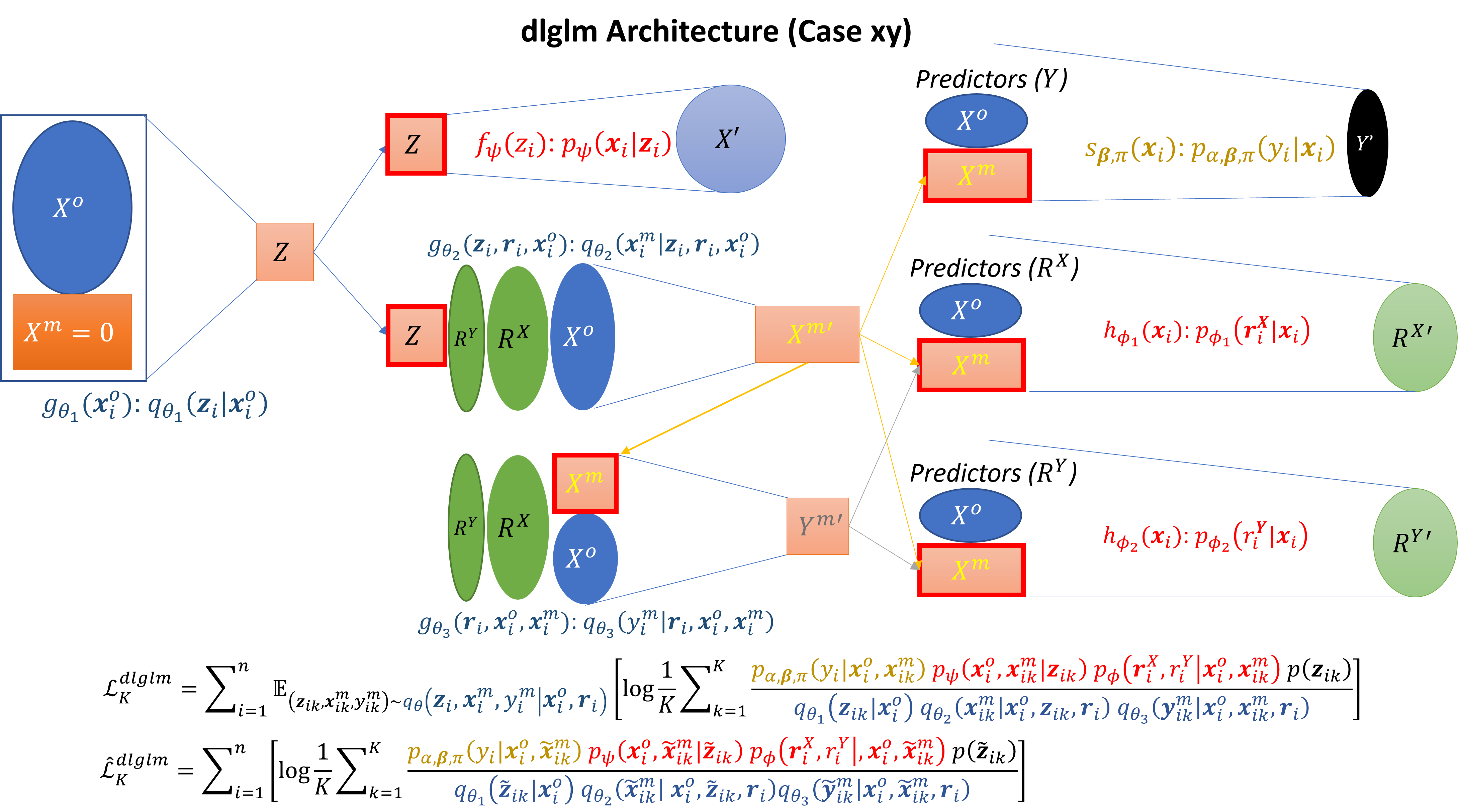} 
\end{center}
\caption{Architecture of proposed dlglm method (\textit{Case xy}). Dark colored nodes ($X^o, X^m=0, R^X, R^Y, Y^o$) represent deterministic values, lightly colored nodes ($Z^{\prime}, X^{o\prime},X^{m\prime}, R^{X\prime}, R^{Y\prime},Y^{\prime}$) represent learned distributional parameters, and outlined (in red) nodes represent sampled values. Orange cells correspond to latent variables $\mathbf{Z}$, $\mathbf{X}^m$, and $\mathbf{Y}^m$. $\mathbf{Z}_1,\ldots, \mathbf{Z}_K$, $\mathbf{X}_{1}^m,\ldots,\mathbf{X}_{K}^m$, and $\mathbf{Y}_{1}^m,\ldots,\mathbf{Y}_{K}^m$ are sampled from their respective variational posteriors $q_{\theta_1}(\mathbf{Z}|\mathbf{X}^o)$, $q_{\theta_2}(\mathbf{X}^m|\mathbf{Z},\mathbf{R},\mathbf{X}^o,\mathbf{Y}^o)$, and $q_{\theta_3}(\mathbf{Y}^m|\mathbf{R},\mathbf{X}^o,\mathbf{X}^m,\mathbf{Y}^o)$. Below is the dlglm bound ($\mathcal{L}^{dlglm}_{K}$), and the estimated dlglm bound ($\hat{\mathcal{L}}^{dlglm}_{K}$), which is optimized via stochastic gradient descent.}
\label{fig:dlglmxyarchitecture}
\end{figure}

In \textit{dlglm}, Case xy is the most general type of missingness that can be encountered, where missingness is present in both the covariates $\mathbf{X}$, as well as the response $\mathbf{Y}$. In this case, one must additionally learn an approximate posterior distribution of the missing response $\mathbf{Y}^m$. The dlglm bound can be derived similarly to the Case x missingness as follows:
\begin{align*}
\log p_{\alpha,\pi,\psi,\phi}(\mathbf{X}^o,\mathbf{Y}^o,\mathbf{R}) &= \sum_{i=1}^{n} \log p_{\alpha,\pi,\psi,\phi}(\mathbf{x}_i^o,y_i^o,\mathbf{r}_i) \\
&= \sum_{i=1}^n \log \left[\iiint p_{\alpha,\pi,\psi,\phi}(\mathbf{x}_i^o,\mathbf{x}_i^m,y_i^o,y_i^m,\mathbf{r}_i^X,r_i^Y,\mathbf{z}_i) d\mathbf{z}_i d\mathbf{x}_i^m dy_i^m \right]\\
&= \sum_{i=1}^n \log \E_{(\mathbf{z}_{ik},\mathbf{x}_{ik}^m,y_{ik}^m) \sim q_{\theta}(\mathbf{z}_i,\mathbf{x}_i^m,y_i^m)} \left[ \frac{1}{K}\sum_{k=1}^K  \frac{p_{\alpha,\pi,\psi,\phi}(\mathbf{x}_i^o,\mathbf{x}_{ik}^m,y_i^o,y_{ik}^m,\mathbf{r}_i^X,r_i^Y,\mathbf{z}_{ik})}{q_\theta(\mathbf{z}_{ik},\mathbf{x}_{ik}^m,y_{ik}^m)} \right]\\
&\geq \sum_{i=1}^{n} \E_{(\mathbf{z}_{ik},\mathbf{x}_{ik}^m,y_{ik}^m) \sim q_{\theta}(\mathbf{z}_i,\mathbf{x}_i^m,y_i^m)} \log{ \left[ \frac{1}{K}\sum_{k=1}^{K}\frac{p_{\alpha,\pi,\psi,\phi}(\mathbf{x}_i^o,\mathbf{x}_{ik}^m,y_i^o,y_{ik}^m,\mathbf{r}_i^X,r_i^Y,\mathbf{z}_{ik})}{q_{\theta}(\mathbf{z}_{ik},\mathbf{x}_{ik}^m,y_{ik}^m)} \right] } \\ &= \mathcal{L}_{K}^{dlglm},
\stepcounter{equation}\tag{\theequation}\label{eqn:dlglmboundxy1}
\end{align*}
where now, we modify the factorization of the joint posterior as $$q_\theta(\mathbf{Z},\mathbf{X}^m,\mathbf{Y}^m)=q_{\theta_1}(\mathbf{Z}|\mathbf{X}^o) q_{\theta_2}(\mathbf{X}^m|\mathbf{Z},\mathbf{X}^o,\mathbf{R}) q_{\theta_3}(\mathbf{Y}^m|\mathbf{X}^o,\mathbf{X}^m,\mathbf{R}).$$ Then, the estimated dlglm bound is
\begin{equation}
\hat{\mathcal{L}}_{K}^{dlglm} = \sum_{i=1}^{n}  \log{\left[\frac{1}{K} \sum_{k=1}^{K} \frac{p_{\alpha,\psi}(y_i|\mathbf{x}_i^o,\tilde{\mathbf{x}}_{ik}^m)p_{\pi}(\mathbf{x}_i|\tilde{\mathbf{z}}_{ik})p(\tilde{\mathbf{z}}_{ik})p_{\phi_1}(\mathbf{r}_i|\mathbf{x}_i^o,\tilde{\mathbf{x}}_{ik}^m)}{q_{\theta_1}(\tilde{\mathbf{z}}_{ik}|\mathbf{x}_i^o) q_{\theta_2}(\tilde{\mathbf{x}}_{ik}^m|\tilde{\mathbf{z}}_{ik},\mathbf{x}_{i}^o,\mathbf{r}_i)q_{\theta_3}(\tilde{y}_{ik}^m|\mathbf{x}_{i}^o,\tilde{\mathbf{x}}_{ik}^m,\mathbf{r}_i)} \right] }. 
\label{eqn:dlglmboundxy2}
\end{equation}
A visualization of the \textit{dlglm} architecture in Case xy can be found in Figure \ref{fig:dlglmxyarchitecture}.

\subsection{A6: Single Imputation via dlglm and idlglm}
Following training, \textit{dlglm} can also perform single imputation of missing values by obtaining point estimates for $\E[\mathbf{x}_i^m|\mathbf{x}_i^o,y_i,\mathbf{r}_i^X]$, defined as the expected value of the missing features given the observed data and the mask for the $i^{th}$ observation under MNAR. We note that imputation is not the main focus of this paper, as it is primarily an unsupervised task, but it may be of interest in some settings to obtain these imputed values for downstream analyses. To do this, we first note that
\begin{align*}
\E[\mathbf{x}_i^m|\mathbf{x}_i^o, y_i, \mathbf{r}_i^X] &= \int \mathbf{x}_i^m p_{\alpha, \boldsymbol{\beta}, \pi, \psi, \phi}(\mathbf{x}_i^m|\mathbf{x}_i^o,y_i,\mathbf{r}_i^X) d\mathbf{x}_i^m \\
&= \iint \mathbf{x}_i^m p_{\alpha, \boldsymbol{\beta}, \pi, \psi, \phi}(\mathbf{x}_i^m,\mathbf{z}_i|\mathbf{x}_i^o,y_i,\mathbf{r}_i^X) d\mathbf{z}_i d\mathbf{x}_i^m \\
&= \iint \mathbf{x}_i^m \dfrac{p_{\alpha, \boldsymbol{\beta}, \pi, \psi, \phi}(\mathbf{x}_i^m,\mathbf{x}_i^o,\mathbf{z}_i,y_i,\mathbf{r}_i^X)}{p_{\psi,\phi}(\mathbf{x}_i^o,y_i,\mathbf{r}_i^X)} d\mathbf{z}_id\mathbf{x}_i^m.
\end{align*}
Then, we can estimate this integral by self-normalized importance sampling. We utilize the proposal density $q_{\theta_1}(\mathbf{z}_i|\mathbf{x}_i^o) q_{\theta_2}(\mathbf{x}_i^m|\mathbf{z}_i,\mathbf{x}_i^o,\mathbf{r}_i^X)$, and define the quantities 
$$w_{ik} = \dfrac{s_{ik}}{s_{i1}+\ldots+s_{iK}} \textrm{, and } s_{ik}=\dfrac{p_{\alpha,\boldsymbol{\beta}, \pi}(y_i|\mathbf{x}_i^o,\tilde{\mathbf{x}}_i^m)p_{\psi}(\mathbf{x}_i|\tilde{\mathbf{z}}_{ik})p(\tilde{\mathbf{z}}_{ik})p_{\phi}(\mathbf{r}_i^X|\mathbf{x}_i^o,\tilde{\mathbf{x}}_{ik}^m)}{q_{\theta_1}(\tilde{\mathbf{z}}_{ik}|\mathbf{x}_i^o) q_{\theta_2}(\tilde{\mathbf{x}}_{ik}^m|\tilde{\mathbf{z}}_{ik},\mathbf{x}_i^o,\mathbf{r}_i^X)}$$
for $k=1,\ldots,K$, with $1$ sample drawn from the variational posterior of each latent variable $\mathbf{z}_{ik}$ and $\mathbf{x}_{ik}^m$ to compute $s_{ik}$, where $w_{ik}$ is defined as standardized ``importance weights'' \citep{Mattei2019}. Using these weights we may estimate $\E[\mathbf{x}_i^m|\mathbf{x}_i^o,y_i,\mathbf{r}_i^X] \approx \sum_{k=1}^{K} w_{ik}\tilde{\mathbf{x}}_{ik}^m$. Then, the process can be repeated for each observation $i=1,\ldots,n$.

In the MCAR or MAR case, one can similarly estimate $\E[\mathbf{x}_i^m|\mathbf{x}_i^o,y_i]$ using the fitted \textit{idlglm} model. By following a similar derivation using the proposal density $q_{\theta_1}(\mathbf{z}_i|\mathbf{x}_i^o)q_{\theta_2}(\mathbf{x}_i^m|\mathbf{z}_i,\mathbf{x}_i^o)$, we obtain the same approximation $\E[\mathbf{x}_i^m|\mathbf{x}_i^o,y_i] \approx \sum_{k=1}^K w_{ik}\tilde{\mathbf{x}}_{ik}^m$, with $w_{ik}$ defined as before, but with a slightly different form for $s_{ik}$: $$ s_{ik}=\dfrac{p_{\alpha,\boldsymbol{\beta}, \pi}(y_i|\mathbf{x}_i^o,\tilde{\mathbf{x}}_i^m)p_{\psi}(\mathbf{x}_i|\tilde{\mathbf{z}}_{ik})p(\tilde{\mathbf{z}}_{ik})}{q_{\theta_1}(\tilde{\mathbf{z}}_{ik}|\mathbf{x}_i^o) q_{\theta_2}(\tilde{\mathbf{x}}_{ik}^m|\tilde{\mathbf{z}}_{ik},\mathbf{x}_i^o)}.$$

\section{Appendix B: Additional Details}
\subsection{B1: Tuning Hyperparameters}
In this section, we summarize the combinations of hyperparameter values that were searched by \textit{dlglm} and \textit{idlglm}. For each dataset, we searched over combinations of two different values for each hyperparameter that was tuned. We tuned 4 different hyperparameters for each method: number of nodes per hidden layer ($h$), number of nodes per hidden layer in the missingness network ($h_r$) , number of hidden layers ($nhl$), number of hidden layers in the $s_{\boldsymbol{\beta},\pi}(\mathbf{x}_i)$ network ($nhl_y$), number of hidden layers in the missingness network ($nhl_r$), dimensionality of latent space ($dz$), and learning rate ($lr$). For \textit{idlglm}, $nhl_r=0$ and $h_r=0$ were fixed (since these are omitted in the \textit{idlglm} architecture). Below are the searched values of the hyperparameters for each dataset in our analyses:

\begin{itemize}
  \item Simulated Data ($p=\{25,50\}$ features)
  \begin{itemize}
    \item $h=\{128,64\}$, $h_r=\{16,32\}$
    \item $lr=\{0.001,0.01\}$
    \item $dz=\{\lfloor 3*p/4 \rfloor, \lfloor p/2 \rfloor, \lfloor p/4 \rfloor, \lfloor p/12 \rfloor \}$
    \item $nhl=\{0,1,2\}$
    \item $nhl_y=0$
    \item $nhl_r=\{0,1\}$
    \item $bs=1,000$, $epochs_{max}=2002$
  \end{itemize}
  \item All UCI (including Banknote) datasets ($p$ features)
  \begin{itemize}
    \item $h=\{128,64\}$, $h_r=\{16,32\}$
    \item $lr=\{0.001,0.01\}$
    \item $dz=\{\lfloor 3*p/4 \rfloor, \lfloor p/2 \rfloor, \lfloor p/4 \rfloor, 8 \}$
    \item $nhl=\{0,1,2\}$
    \item $nhl_y=\{0,1,2\}$
    \item $nhl_r=\{0,1\}$
    \item $bs=1000$, $epochs_{max}=2002$
  \end{itemize}
\end{itemize}

\subsection{B2: Additional Simulations}
\begin{figure}
\begin{center}
\includegraphics[width=165mm]{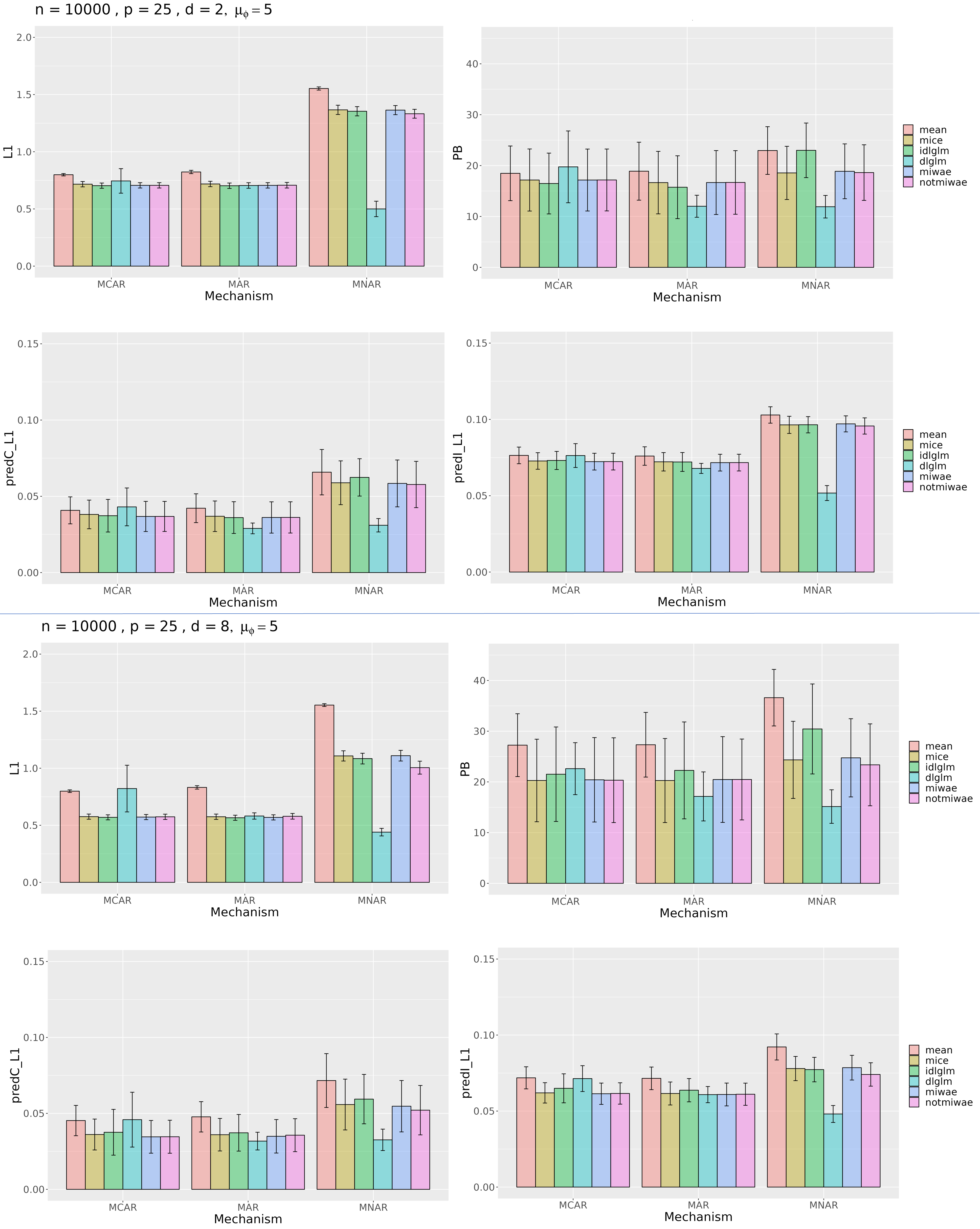}
\end{center}
\caption{\scriptsize Simulation results with $n=10,000$ and $p=25$, varying $d=2$ (top 4) and $d=8$ (bottom 4). In each quadrant, we measure imputation accuracy by the average L1 distance between imputed vs true values in $\mathbf{X}$ (top-left), coefficient estimation accuracy by the average percent bias (PB) of the estimates $\hat{\boldsymbol{\beta}}$ compared to the truth (top-right), and prediction accuracy by the average L1 distance between the predicted and true probabilities of class 1 membership of $\mathbf{Y}$ using the true unmasked test set (predC, bottom-left) and the incomplete test set (predI, bottom-right). In predI, we first impute missing values of the test data for mean, \textit{miwae}, \textit{notmiwae}, and \textit{mice} imputation, and we input the incomplete test set as is for \textit{dlglm} and \textit{idlglm}.}
\label{fig:dlglmsimsN10KP25}
\end{figure}

\begin{figure}
\begin{center}
\includegraphics[width=165mm]{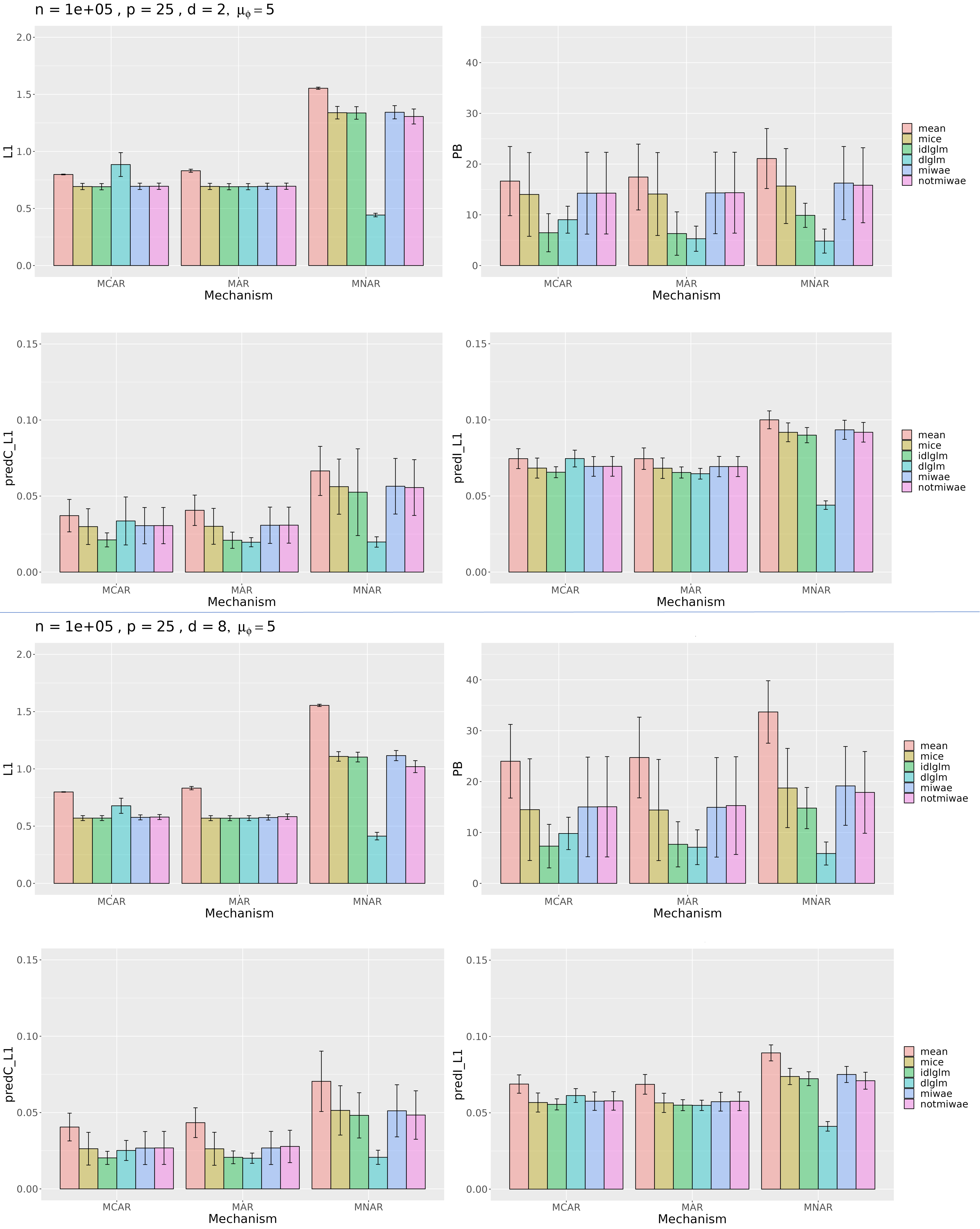}
\end{center}
\caption{\scriptsize Simulation results with $n=100,000$ and $p=25$, varying $d=2$ (top 4) and $d=8$ (bottom 4). In each quadrant, we measure imputation accuracy by the average L1 distance between imputed vs true values in $\mathbf{X}$ (top-left), coefficient estimation accuracy by the average percent bias (PB) of the estimates $\hat{\boldsymbol{\beta}}$ compared to the truth (top-right), and prediction accuracy by the average L1 distance between the predicted and true probabilities of class 1 membership of $\mathbf{Y}$ using the true unmasked test set (predC, bottom-left) and the incomplete test set (predI, bottom-right). In predI, we first impute missing values of the test data for mean, \textit{miwae}, \textit{notmiwae}, and \textit{mice} imputation, and we input the incomplete test set as is for \textit{dlglm} and \textit{idlglm}.}
\label{fig:dlglmsimsN100KP25}
\end{figure}

Supplementary Figures \ref{fig:dlglmsimsN10KP25} and \ref{fig:dlglmsimsN100KP25} show the results of the imputation, coefficient estimation, and prediction performance on simulated data with $p=25$ fixed, with the default (linear) simulation of missingness.

\begin{figure}
\begin{center}
\includegraphics[width=165mm]{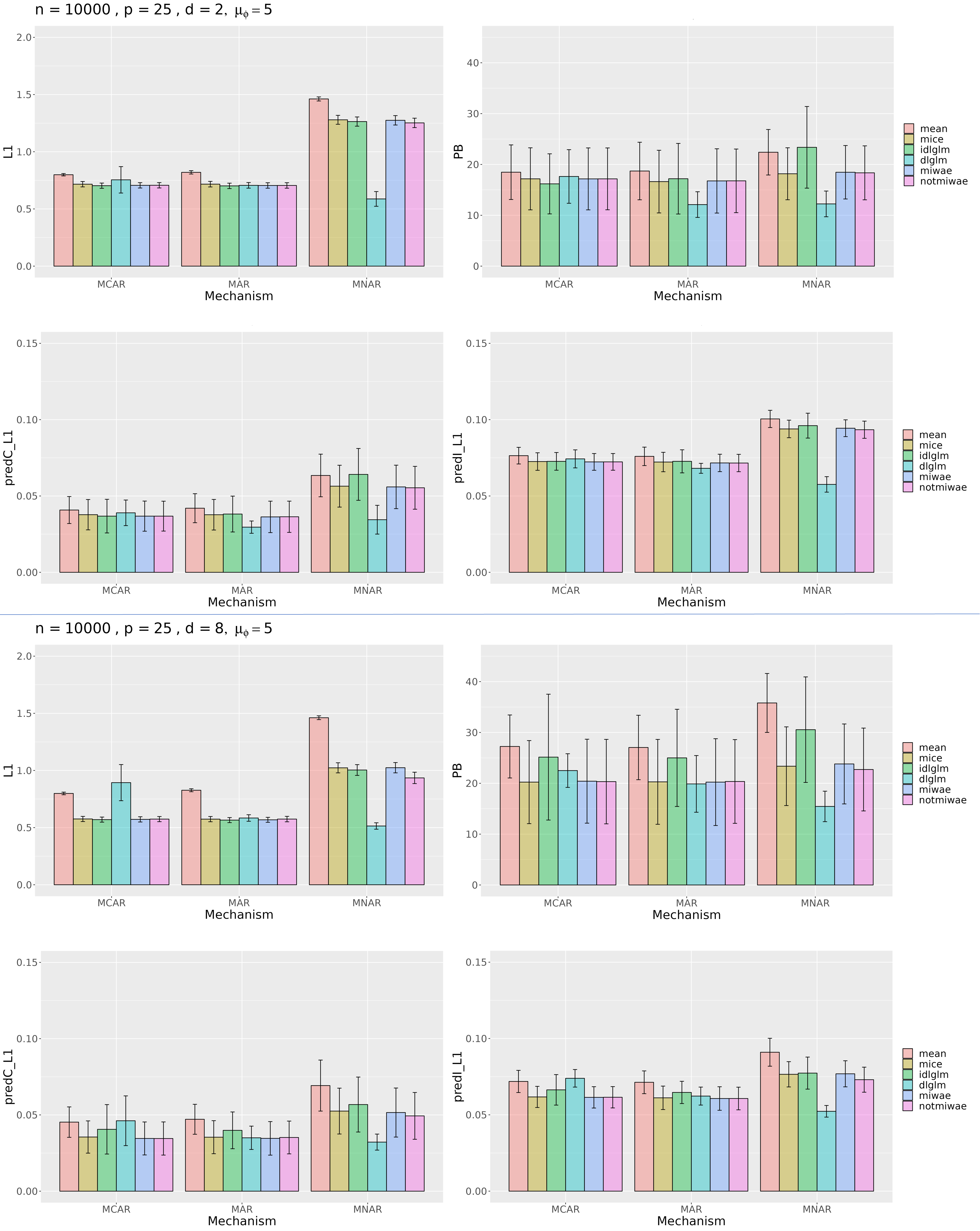}
\end{center}
\caption{\scriptsize Simulation results with $n=100,000$ and $p=25$, varying $d=2$ (top 4) and $d=8$ (bottom 4), with nonlinear generation of the missingness mask. In each quadrant, we measure imputation accuracy by the average L1 distance between imputed vs true values in $\mathbf{X}$ (top-left), coefficient estimation accuracy by the average percent bias (PB) of the estimates $\hat{\boldsymbol{\beta}}$ compared to the truth (top-right), and prediction accuracy by the average L1 distance between the predicted and true probabilities of class 1 membership of $\mathbf{Y}$ using the true unmasked test set (predC, bottom-left) and the incomplete test set (predI, bottom-right). In predI, we first impute missing values of the test data for mean, \textit{miwae}, \textit{notmiwae}, and \textit{mice} imputation, and we input the incomplete test set as is for \textit{dlglm} and \textit{idlglm}.}
\label{fig:dlglmNLsimsN10KP25}
\end{figure}

\begin{figure}
\begin{center}
\includegraphics[width=165mm]{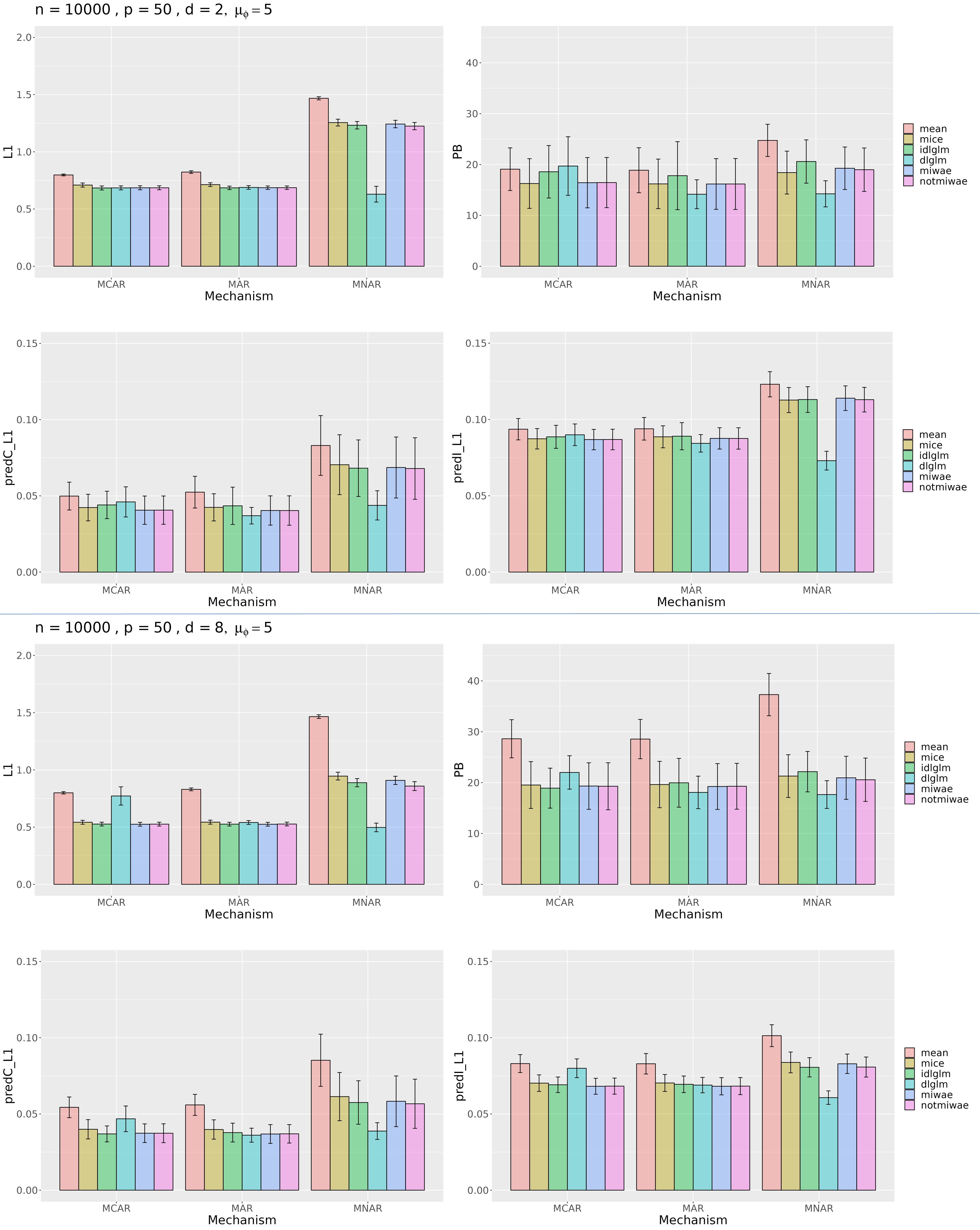}
\end{center}
\caption{\scriptsize Simulation results with $n=100,000$ and $p=25$, varying $d=2$ (top 4) and $d=8$ (bottom 4), with nonlinear generation of the missingness mask. In each quadrant, we measure imputation accuracy by the average L1 distance between imputed vs true values in $\mathbf{X}$ (top-left), coefficient estimation accuracy by the average percent bias (PB) of the estimates $\hat{\boldsymbol{\beta}}$ compared to the truth (top-right), and prediction accuracy by the average L1 distance between the predicted and true probabilities of class 1 membership of $\mathbf{Y}$ using the true unmasked test set (predC, bottom-left) and the incomplete test set (predI, bottom-right). In predI, we first impute missing values of the test data for mean, \textit{miwae}, \textit{notmiwae}, and \textit{mice} imputation, and we input the incomplete test set as is for \textit{dlglm} and \textit{idlglm}.}
\label{fig:dlglmNLsimsN10KP50}
\end{figure}

\begin{figure}
\begin{center}
\includegraphics[width=165mm]{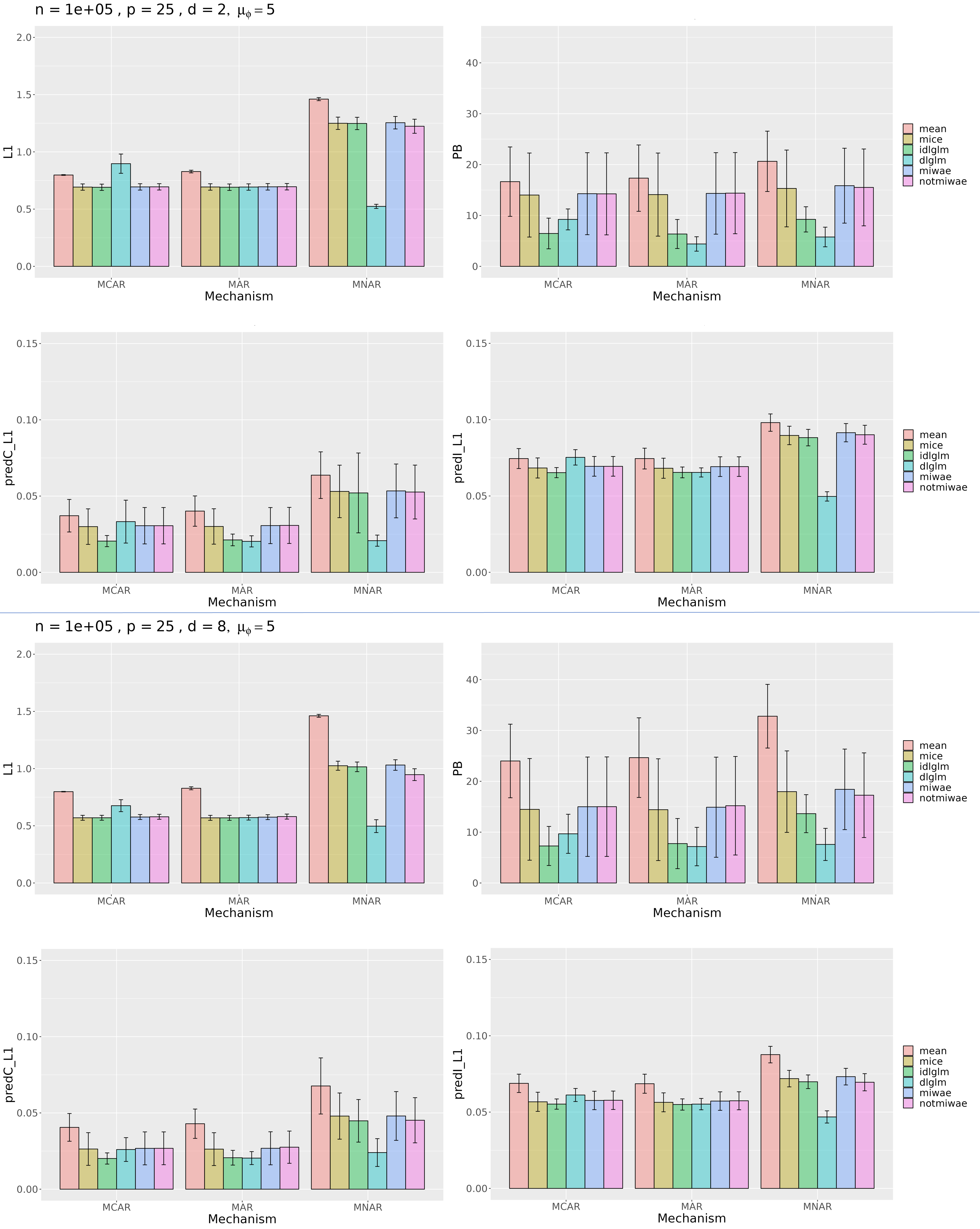}
\end{center}
\caption{\scriptsize Simulation results with $n=100,000$ and $p=25$, varying $d=2$ (top 4) and $d=8$ (bottom 4), with nonlinear generation of the missingness mask. In each quadrant, we measure imputation accuracy by the average L1 distance between imputed vs true values in $\mathbf{X}$ (top-left), coefficient estimation accuracy by the average percent bias (PB) of the estimates $\hat{\boldsymbol{\beta}}$ compared to the truth (top-right), and prediction accuracy by the average L1 distance between the predicted and true probabilities of class 1 membership of $\mathbf{Y}$ using the true unmasked test set (predC, bottom-left) and the incomplete test set (predI, bottom-right). In predI, we first impute missing values of the test data for mean, \textit{miwae}, \textit{notmiwae}, and \textit{mice} imputation, and we input the incomplete test set as is for \textit{dlglm} and \textit{idlglm}.}
\label{fig:dlglmNLsimsN100KP25}
\end{figure}

Supplementary Figures \ref{fig:dlglmNLsimsN10KP25} - \ref{fig:dlglmNLsimsN100KP25} show these results on simulated data with $n=10,000$ with $p=\{25,50\}$, and $n=100,000$ with $p=25$.

\begin{figure}
\begin{center}
\includegraphics[width=165mm]{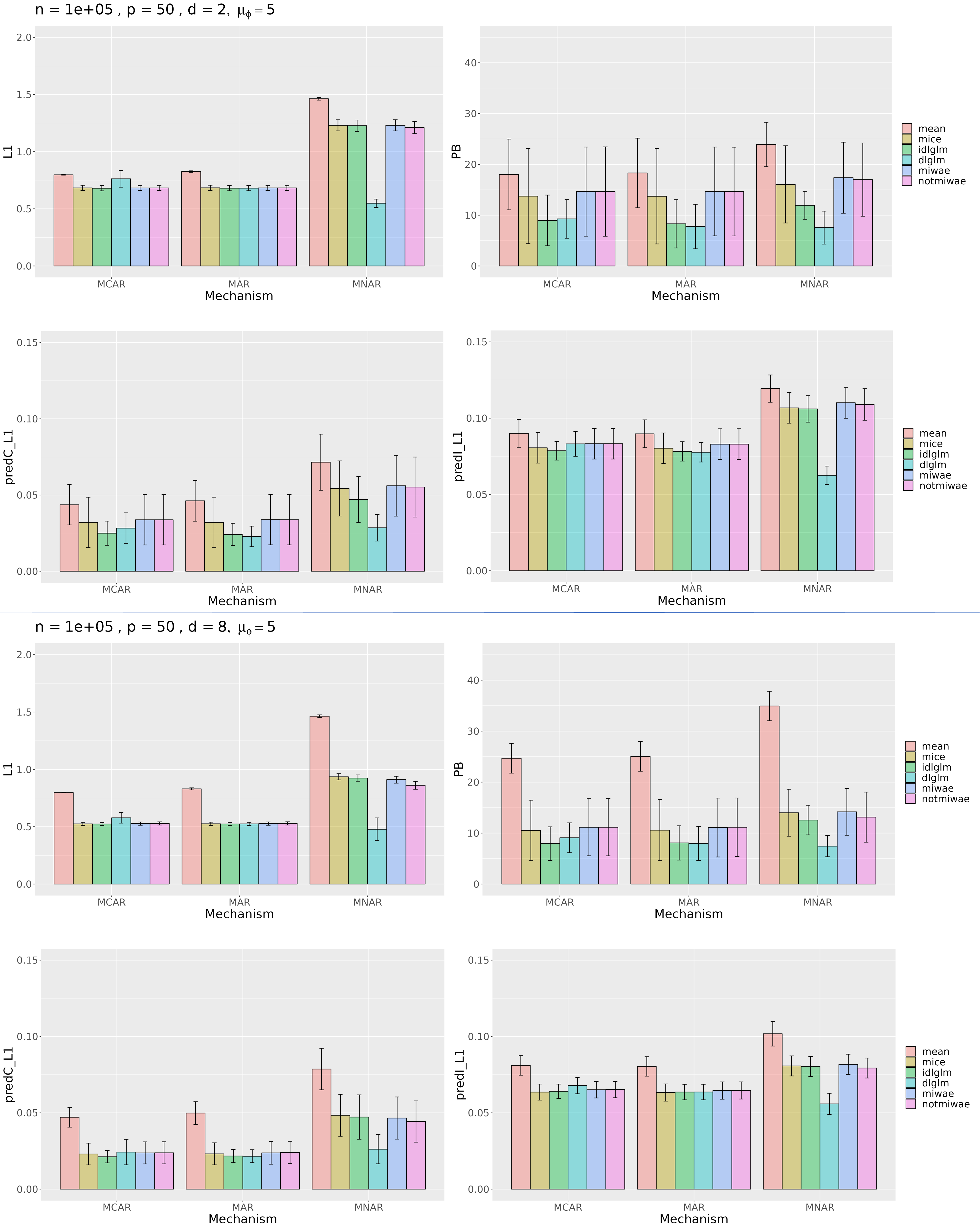}
\end{center}
\caption{\scriptsize Simulation results with $n=100,000$ and $p=50$, varying $d=2$ (top 4) and $d=8$ (bottom 4), with nonlinear generation of the missingness mask. In each quadrant, we measure imputation accuracy by the average L1 distance between imputed vs true values in $\mathbf{X}$ (top-left), coefficient estimation accuracy by the average percent bias (PB) of the estimates $\hat{\boldsymbol{\beta}}$ compared to the truth (top-right), and prediction accuracy by the average L1 distance between the predicted and true probabilities of class 1 membership of $\mathbf{Y}$ using the true unmasked test set (predC, bottom-left) and the incomplete test set (predI, bottom-right). In predI, we first impute missing values of the test data for mean, \textit{miwae}, \textit{notmiwae}, and \textit{mice} imputation, and we input the incomplete test set as is for \textit{dlglm} and \textit{idlglm}.}
\label{fig:dlglmNLsimsN100KP50}
\end{figure}

Next, we show the results of the ``nonlinear missingness" simulation cases in Figures \ref{fig:dlglmNLsimsN10KP25}-\ref{fig:dlglmNLsimsN100KP50}. We note here that the MCAR case represents the same generation of missingness as before, but we include it here for side-by-side comparisons. Interestingly, the compared methods performed very similarly under MAR, since the complexity of the underlying missingness mechanism for ignorably-missing data. Under MNAR, we see a slight increase in the imputation L1, percent bias, and prediction error metrics for \textit{dlglm}, since the underlying mechanism of missingness is more complex. However, the metrics are still comparable to those from the previously used missingness model, showing robustness to complex nonlinearity in the missingness model. Additionally, \textit{dlglm} still consistently outperforms competing methods under MNAR missingness.

\section{Appendix C: Data}
Datasets used for analysis in this study are all publicly available online. Details of how to obtain these datasets are given in this section.
\subsection{UCI Machine Learning Datasets}
The UCI datasets can be obtained from the UCI Machine Learning repository at \url{https://archive.ics.uci.edu/ml/index.php} \citep{Dua2017} The datasets included in the analyses in this study (and in Appendix D) are given below:
\begin{enumerate}
  \item CAREVALUATION: $n=1728$ and $p=6$.
    \begin{itemize}
      \item Documentation: \url{https://archive.ics.uci.edu/ml/datasets/car+evaluation}
      \item Link to Data: \url{https://archive.ics.uci.edu/ml/machine-learning-databases/car/car.data}
      \item Response of interest: Class variable (last feature)
    \end{itemize}
  \item BANKNOTE: $n=1372$ and $p=4$.
    \begin{itemize}
      \item Documentation: \url{https://archive.ics.uci.edu/ml/datasets/banknote+authentication}
      \item Link to Data: \url{https://archive.ics.uci.edu/ml/machine-learning-databases/00267/data_banknote_authentication.txt}
      \item Response of interest: Class variable (last feature)
    \end{itemize}
  \item DRYBEAN: $n=13611$ and $p=16$
    \begin{itemize}
        \item Documentation: \url{https://archive.ics.uci.edu/ml/datasets/Dry+Bean+Dataset}
        \item Link to Data: \url{https://archive.ics.uci.edu/ml/machine-learning-databases/00602/DryBeanDataset.zip}
        \item Response of interest: Class variable (last feature)
    \end{itemize}
  \item LETTER: $n=20000$ and $p=16$
    \begin{itemize}
        \item Documentation: \url{https://archive.ics.uci.edu/ml/datasets/letter+recognition}
        \item Link to Data: \url{https://archive.ics.uci.edu/ml/machine-learning-databases/letter-recognition/letter-recognition.data}
        \item Response of interest: Class variable (first feature)
    \end{itemize}
  \item RED: $n=1599$ and $p=11$
    \begin{itemize}
      \item Documentation: \url{https://archive.ics.uci.edu/ml/datasets/wine+quality}
      \item Link to Data: \url{https://archive.ics.uci.edu/ml/machine-learning-databases/wine-quality/winequality-red.csv}
      \item Response of interest: Wine quality score (last feature)
    \end{itemize}
  \item SHUTTLE: $n=57756$ (after pre-filtering low-incidence classes) and $p=9$
    \begin{itemize}
        \item Documentation: \url{https://archive.ics.uci.edu/ml/datasets/Statlog+(Shuttle)}
        \item Link to Data: Attained using the \textit{mlbench} R package, using the command: \verb|data(Shuttle)|
        \item Response of interest: Class variable (last feature)
    \end{itemize}
  \item SPAM: $n=4601$ and $p=57$
    \begin{itemize}
        \item Documentation: \url{https://archive.ics.uci.edu/ml/datasets/spambase}
        \item Link to Data: \url{https://archive.ics.uci.edu/ml/machine-learning-databases/spambase/spambase.data}
        \item Response of interest: Class variable (last feature)
    \end{itemize}
  \item WHITE: $n=4898$ and $p=11$
    \begin{itemize}
      \item Documentation: \url{https://archive.ics.uci.edu/ml/datasets/wine+quality}
      \item Link to Data: \url{https://archive.ics.uci.edu/ml/machine-learning-databases/wine-quality/winequality-white.csv}
      \item Response of interest: Wine quality score (last feature)
    \end{itemize}
\end{enumerate}

\begin{figure}
\begin{center}
\includegraphics[width=165mm]{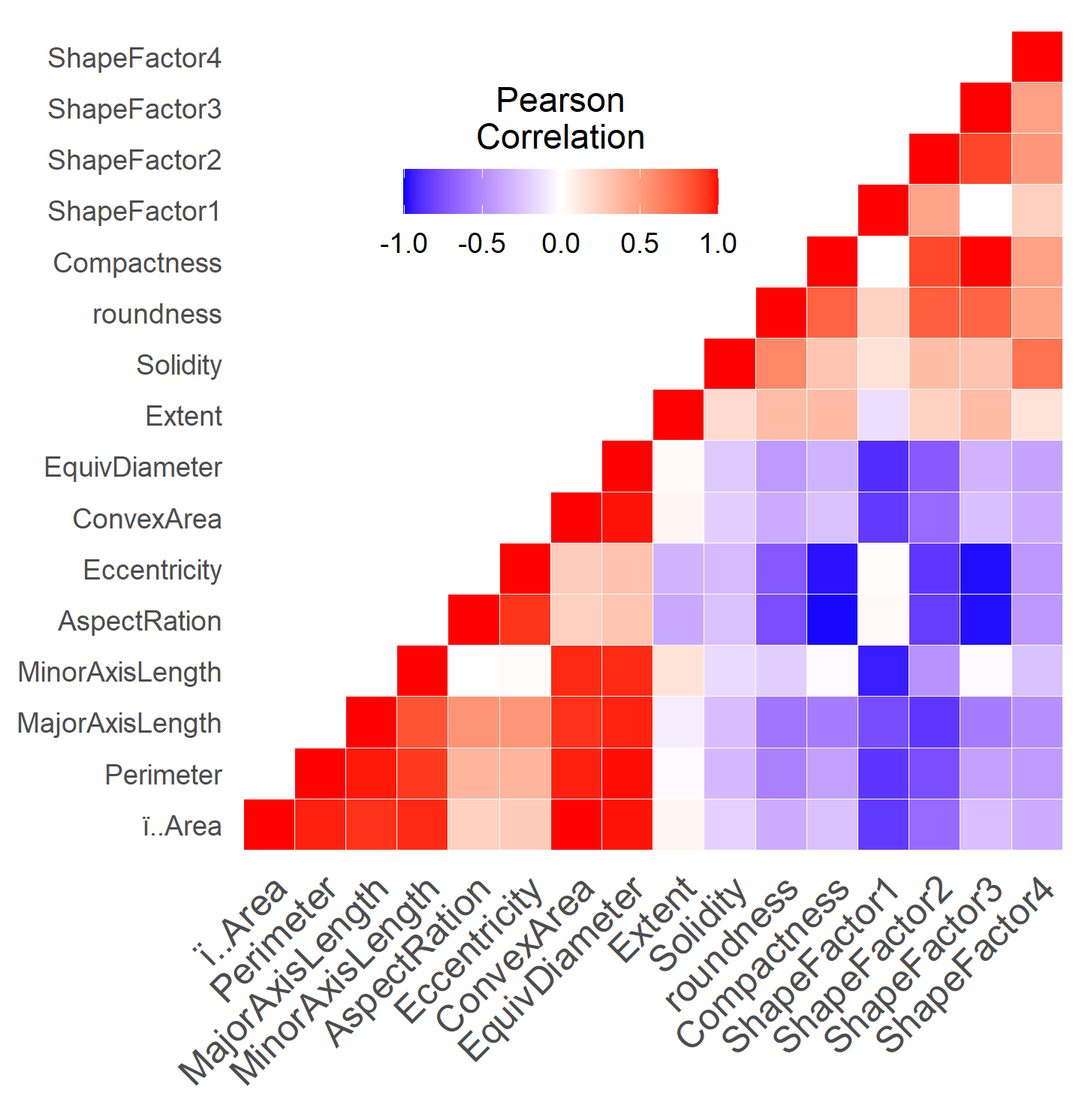}
\end{center}
\caption{\scriptsize Correlation heatmap matrix of features in the DRYBEAN dataset. Many features in this dataset were highly correlated.}
\label{fig:drybeancormat}
\end{figure}

In particular, the DRYBEAN dataset contained very highly correlated features, as can be seen in the heatmap in Supplementary Figure \ref{fig:drybeancormat}. Such high correlation between features may allow ignorably-missing methods do perform well under non-ignorable missingness, as the partially-observed features that the missingness depends on may be highly correlated with completely-observed features in the dataset.

\section{Appendix D: Additional Simulations}

\begin{sidewaysfigure}
\begin{center}
\includegraphics[width=205mm]{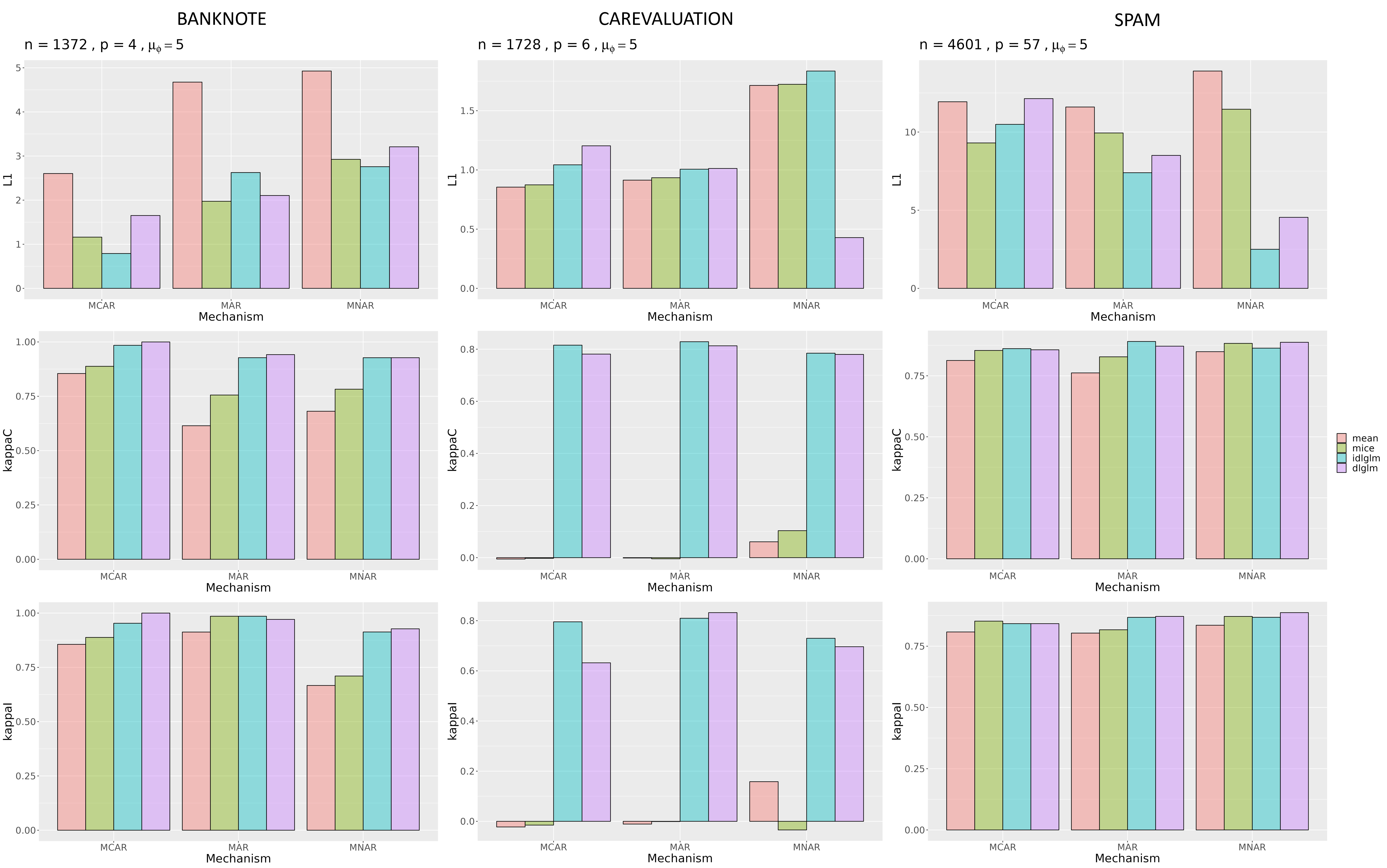}
\end{center}
\caption{\scriptsize Imputation (top row) and prediction results from predC (middle row) and predI (bottom row) from comparative methods run on 3 small datasets from the UCI Machine Learning Repository: BANKNOTE, CAREVALUATION, and SPAM (columns, left to right). Imputation error was measured by the average L1 distance between true and imputed entries, with lower values indicating better performance, and prediction performance was measured by the Cohen's kappa metric for both predC (kappaC) and predI (kappaI), with higher values indicating better performance.}
\label{fig:dlglmUCI2}
\end{sidewaysfigure}

\begin{sidewaysfigure}
\begin{center}
\includegraphics[width=205mm]{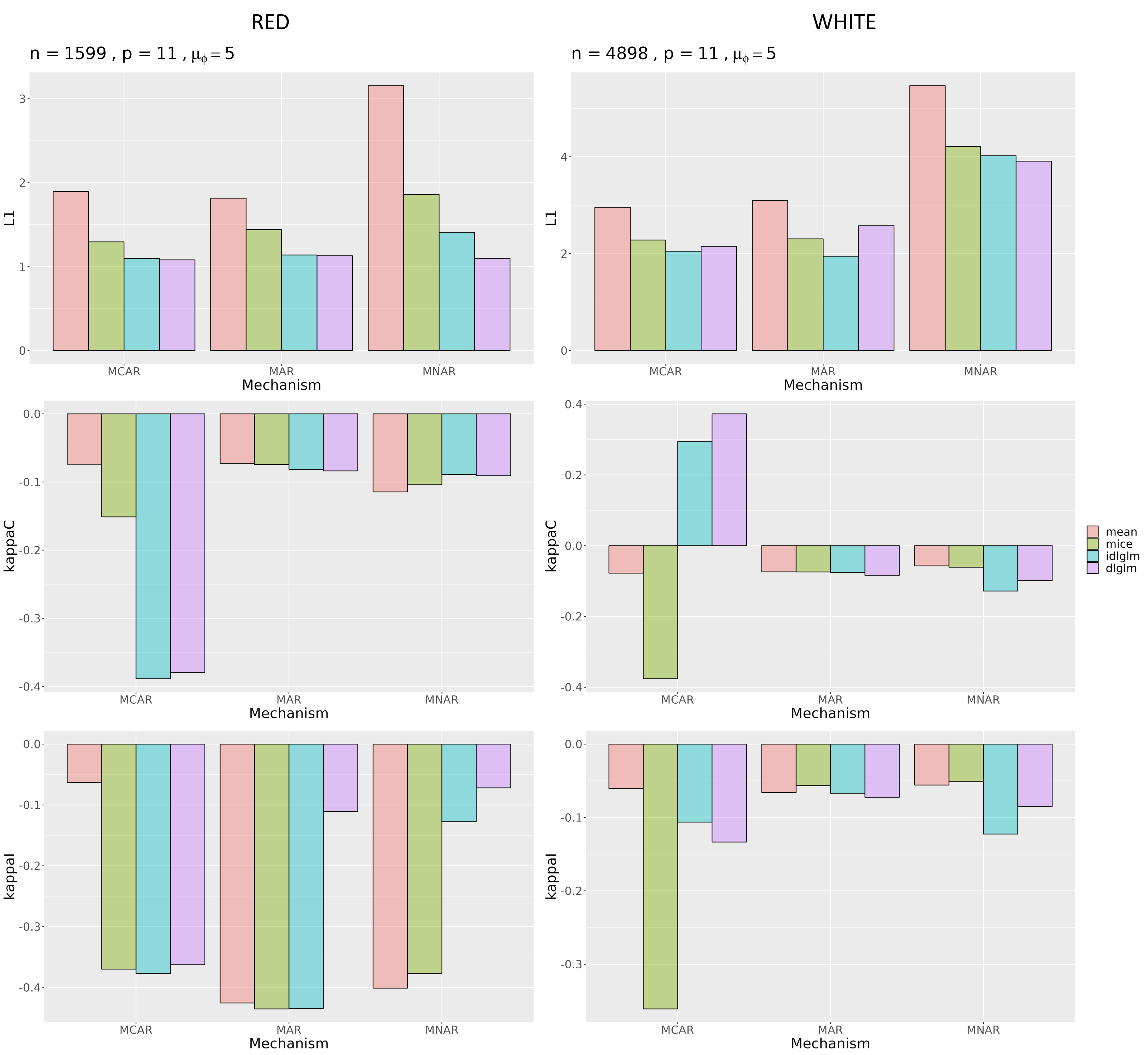}
\end{center}
\caption{\scriptsize Imputation (top row) and prediction results from predC (middle row) and predI (bottom row) from comparative methods run on 2 small datasets from the UCI Machine Learning Repository: RED and WHITE (columns, left and right). Imputation error was measured by the average L1 distance between true and imputed entries, with lower values indicating better performance, and prediction performance was measured by the Cohen's kappa metric for both predC (kappaC) and predI (kappaI), with higher values indicating better performance.}
\label{fig:dlglmUCI3}
\end{sidewaysfigure}

We also performed comparative analyses in 5 other small UCI datasets, as in Section 3.2 in the main text. Supplementary Figures \ref{fig:dlglmUCI2} and \ref{fig:dlglmUCI3} show the results for these analyses for the BANKNOTE, CAREVALUATION, SPAM, RED, and WHITE datasets. In general, we see that \textit{dlglm} still performs best in imputation and prediction under MNAR missingness, although \textit{idlglm} performs slightly better in the BANKNOTE dataset. In lower sample size settings, \textit{dlglm} may not perform optimally, due to the more complex architecture and the additional task of learning a missingness model. We also note that for these missingness models in \textit{dlglm}, as in the main text, we included all of the features in \textbf{X} as well as the response variable \textbf{Y} as covariates, although only one feature in \textbf{X} is involved in the simulation of missingness in each incomplete feature. Selecting important features from an overparameterized missingness model may be very difficult, and thus we recommend \textit{dlglm} to be used with a sample size of at least $n=10000$.

\section{Appendix E: Bank Marketing Dataset}
The Bank Marketing dataset contains 17 attributes collected from 41,188 clients from direct marketing campaigns of a Portuguese banking institution. The dataset is publicly available here: \url{https://archive.ics.uci.edu/ml/machine-learning-databases/00222/bank-additional.zip}. Any ``unknown'' or ``nonexistent'' values in the data were deemed to be missing. Also, a value of 999 for $pdays$ (the number of days that passed by after the client was last contacted from a previous campaign) was denoted as missing. The response variable of interest was a binary indicator of whether the client subscribed to the term deposit. We omitted the month and day of the call due to irrelevance with respect to the response variable. We also removed the call duration variable as stated in the documentation, due to a strong correlation with the response. The resulting dataset contained 16 features for each of the 41,188 clients. More information about this dataset can be found in \url{https://archive.ics.uci.edu/ml/datasets/bank+marketing}.

\section{Appendix F: Additional Metrics}
The formula for the PPV and F1 metrics are given as
\begin{align*}
    PPV &= \frac{TP}{TP+TN}\\
    F1 &= \frac{2*TP}{2*TP+FP+FN},
\end{align*}
where TP, TN, FP, and FN are the number of True Positives, True Negatives, False Positives, and False Negatives, respectively. In this dataset, a ``positive" denotes a client that subscribed to the term deposit, while a ``negative" denotes a client that did not subscribe.

\bibliographystyle{agsm}
\bibliography{dlglm_JCGS}